\documentclass[journal]{IEEEtran}
\usepackage{amsmath,amsfonts}
\usepackage{latexsym}
\usepackage{algorithmic}
\usepackage{algorithm}
\usepackage{array}
\usepackage[caption=false,font=normalsize,labelfont=sf,textfont=sf]{subfig}
\usepackage{textcomp}
\usepackage{times}
\usepackage{epsfig}
\usepackage{cuted}
\usepackage{stfloats}
\usepackage{url}
\usepackage{booktabs} 
\usepackage{verbatim}
\usepackage{graphicx}
\usepackage[export]{adjustbox}
\usepackage{amsmath,amssymb} 
\usepackage{multirow}
\usepackage{rotating}
\usepackage{wrapfig}
\usepackage{float}

\usepackage{amssymb}

\usepackage[ruled,vlined,algo2e]{algorithm2e} 

\usepackage{multirow}
\usepackage{caption}
\usepackage[colorlinks,linkcolor=blue]{hyperref}
\newcommand{\onedot}{\ifx\@let@token.\else.\null\fi\xspace}
\newcommand{\etal}{\emph{et al}\onedot}
\newcommand{\eg}{\emph{e.g}\onedot}
\newcommand{\ie}{\emph{i.e}\onedot}

\usepackage{color}
\definecolor{yfcolor}{RGB}{255,0,0}
\newcommand{\yf}[1]{{\color{yfcolor}#1}}

\usepackage{xcolor}
\newcommand{\re}[1]{{\color{black} #1}}
\newcommand{\red}[1]{{\color{black} #1}}
\newcommand{\blue}[1]{{\color{black} #1}}
\begin{document}

\title{On Robust Cross-View Consistency in Self-Supervised Monocular Depth Estimation}

\author{Haimei Zhao, Jing Zhang, Zhuo Chen, Bo Yuan, ~\IEEEmembership{Senior Member,~IEEE}, Dacheng Tao, ~\IEEEmembership{Fellow,~IEEE}
\thanks{\textit{Corresponding author: Dacheng Tao.}}
\thanks{H. Zhao, J. Zhang and D. Tao are with the School of Computer Science, Faculty of Engineering, The University of Sydney, 6 Cleveland St, Darlington, NSW 2008, Australia (email: hzha7798@uni.sydney.edu.au; jing.zhang1@sydney.edu.au; dacheng.tao@gmail.com).}
\thanks{Z. Chen and Bo Yuan are with the Shenzhen International Graduate School, Tsinghua University, Shenzhen 518055, China (e-mail: z-chen17@mails.tsinghua.edu.cn; boyuan@ieee.org).}}


 \markboth{Submitted to Journal}%
{Shell \MakeLowercase{\textit{et al.}}: A Sample Article Using IEEEtran.cls for IEEE Journals}


\maketitle

\begin{abstract}
Remarkable progress has been made in self-supervised monocular depth estimation (SS-MDE) by exploring cross-view consistency, e.g., photometric consistency and 3D point cloud consistency. However, they are very vulnerable to illumination variance, occlusions, texture-less regions, as well as moving objects, making them not robust enough to deal with various scenes. To address this challenge, we study two kinds of robust cross-view consistency in this paper. Firstly, the spatial offset field between adjacent frames is obtained by reconstructing the reference frame from its neighbors via deformable alignment, which is used to align the temporal depth features via a Depth Feature Alignment (DFA) loss. Secondly, the 3D point clouds of each reference frame and its nearby frames are calculated and transformed into voxel space, where the point density in each voxel is calculated and aligned via a Voxel Density Alignment (VDA) loss. In this way, we exploit the temporal coherence in both depth feature space and 3D voxel space for SS-MDE, shifting the ``point-to-point'' alignment paradigm to the ``region-to-region'' one. Compared with the photometric consistency loss as well as the rigid point cloud alignment loss, the proposed DFA and VDA losses are more robust owing to the strong representation power of deep features as well as the high tolerance of voxel density to the aforementioned challenges. Experimental results on \re{several outdoor benchmarks} show that our method outperforms current state-of-the-art techniques. Extensive ablation study and analysis validate the effectiveness of the proposed losses, especially in challenging scenes. \re{The code and models are available at}~\href{https://github.com/sunnyHelen/RCVC-depth}{https://github.com/sunnyHelen/RCVC-depth}.
\end{abstract}


\section{Introduction}
Understanding the 3D structure of scenes is an essential topic in machine perception, which plays a crucial part in autonomous driving and robot vision. Traditionally, this task can be accomplished by Structure from Motion and with multi-view or binocular stereo inputs \cite{bjorkman2002real}.
Since stereo images are more expensive and inconvenient to acquire than monocular ones, solutions based on monocular vision have attracted increasing attention from the community. However, monocular depth estimation is generally more challenging than stereo methods due to scale ambiguity and unknown camera motion. Several works \cite{eigen2014depth, godard2017unsupervised} have been proposed to narrow the performance gap.
\begin{figure}[htbp]
  \begin{center}
   \includegraphics[width=0.48\textwidth]{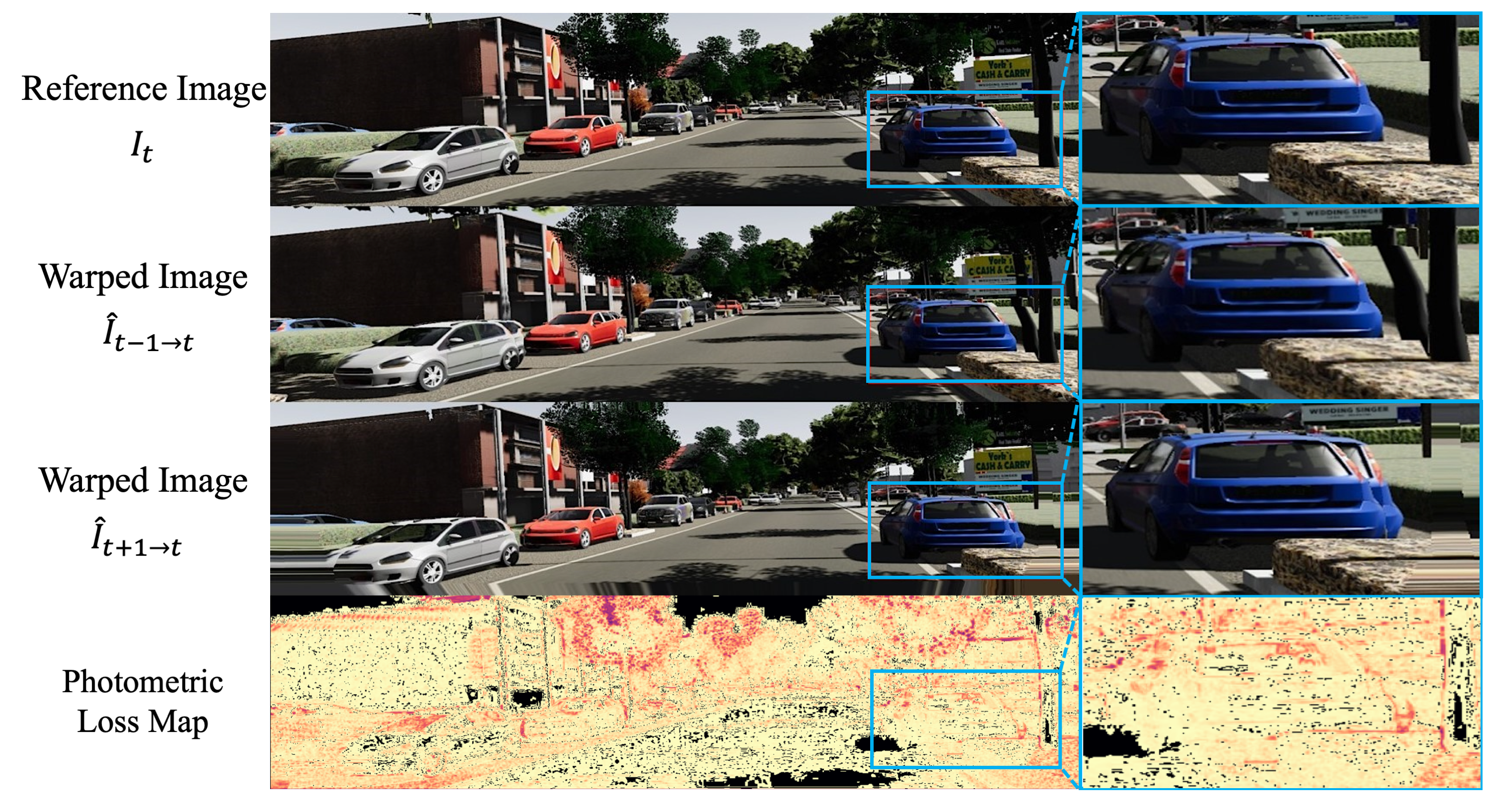}
  \end{center}
  \caption{Visualization of the photometric loss. {(Black areas represent regions without photometric loss, whereas warmer colors indicate larger loss values in other regions.)}. The first row is the reference image, and the second and third rows are warped images from adjacent images using ground truth depth and pose.}
\label{fig:photometric}
\end{figure}
\begin{figure*}
\begin{center}
\begin{minipage}[htbp]{\textwidth}
\centering
\includegraphics[width=14.2cm]{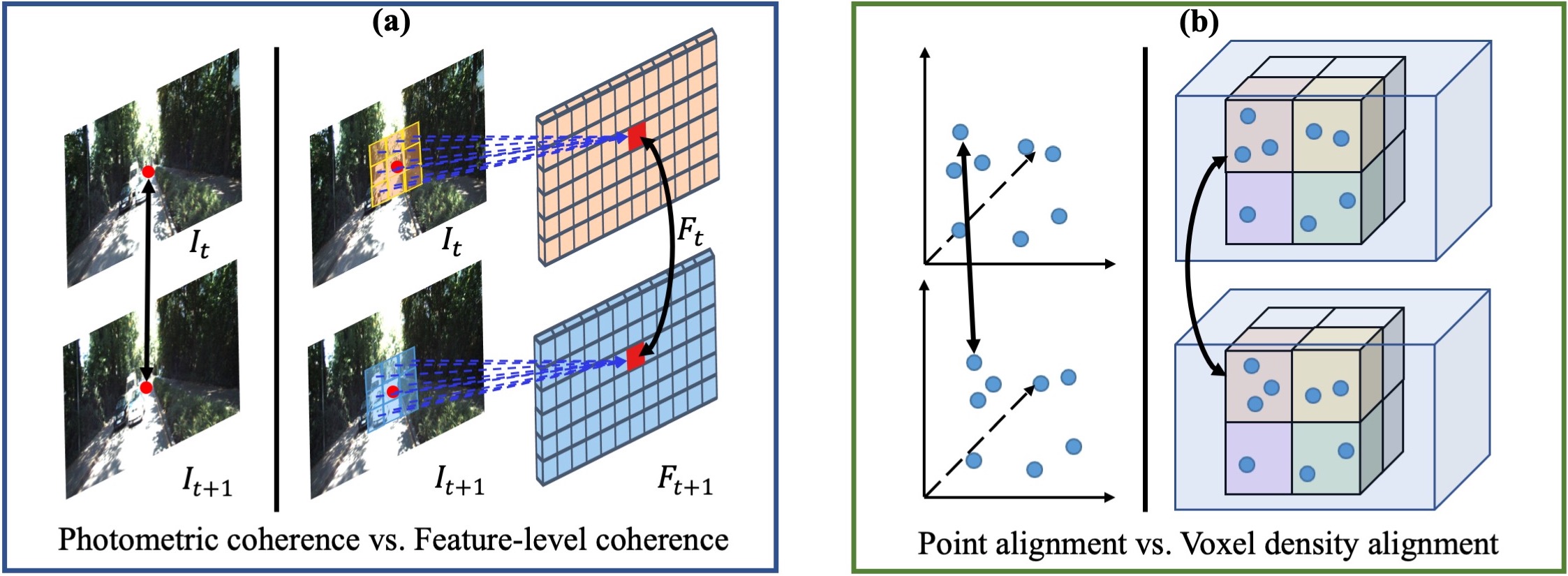}
\end{minipage}
\end{center}
\caption{Comparisons of prior ``point-to-point" alignment paradigm to our ``region-to-region" one. \re{We propose the ``region-to-region" alignment paradigm by enforcing photometric consistency at feature-level (a) and replacing point cloud alignment with voxel density alignment in 3D space (b).}}
\label{fig:1}
\end{figure*}

Recently, with the unprecedented success of deep learning in computer vision \cite{dosovitskiy2020image,jing2021amalgamating,mo2021terra,xi2022differentiable}, Convolutional Neural Networks (CNNs) \cite{he2016deep} have achieved promising results in the field of depth estimation. In the paradigm of supervised learning, depth estimation is usually regarded as a regression or classification problem \cite{eigen2014depth,fu2018deep}, which needs expensive labeled datasets. 
By contrast, there are also some successful attempts \cite{zhou2017unsupervised,mahjourian2018unsupervised,godard2019digging,zhao2020collaborative,zhao2022jperceiver,zhang2022towards1,zhang2022towards2} to execute monocular depth estimation and visual odometry prediction together in a self-supervised manner by utilizing cross-view consistency between consecutive frames. In most prior works of this pipeline, two networks are used to predict the depth and the camera pose separately, which are then jointly exploited to warp source frames to the reference ones, thereby converting the depth estimation problem to a photometric error minimization process. The essence of this paradigm is utilizing the cross-view geometry consistency from videos to regularize the joint learning of depth and pose.

Previous SS-MDE works have proved the effectiveness of the photometric loss among consecutive frames, but it is quite vulnerable and even problematic in some cases. First, the photometric consistency is based on the assumption that the pixel intensities projected from the same 3D point in different frames are constant, which is easily violated by illumination variance, reflective surface, and texture-less region. Second, there are always some dynamic objects in natural scenes and thus generating occlusion areas, which also affects the success of photometric consistency. To demonstrate the vulnerability of photometric loss, we conduct a preliminary study on Virtual KITTI \cite{cabon2020virtual} because it has dense ground truth depth maps and precise poses.
As shown in Figure \ref{fig:photometric}, even though the ground truth depth and pose are used, the photometric loss map is always not zero due to factors such as occlusions, illumination variance, dynamic objects, etc. To address this problem, the Perceptual losses are used in recent work \cite{shu2020featdepth}. In line with this research direction, we are dedicated to proposing more robust loss items to help enhance the self-supervision signal.

Therefore, our work targets to explore more robust cross-view consistency losses to mitigate the side effect of these challenging cases. We first propose a Depth Feature Alignment (DFA) loss, which learns feature offsets between consecutive frames by reconstructing the reference frames from its adjacent frames via deformable alignment. Then, these feature offsets are used to align the temporal depth feature. In this way, we utilize the consistency between adjacent frames via feature-level representation, which is more representative and discriminative than pixel intensities. 
As shown in Figure \ref{fig:1} (a), comparing the photometric intensity between consecutive frames can be problematic, because the intensities of the surrounding region of the target pixel are very close, and the ambiguity may probably cause mismatches. 

Besides, prior work \cite{mahjourian2018unsupervised}
proposes to use ICP-based point cloud alignment loss to utilize 3D geometry to enforce cross-view consistency, which is useful to alleviate the ambiguity of 2D pixels. However, rigid 3D point cloud alignment can not work properly in scenes with the object motion and the resulting occlusion, as shown in Figure \ref{fig:1} (b), thereby being sensitive to local object motion. In order to make the model more robust to moving objects and the resulting occlusion areas, we propose voxel density as a new 3D representation and define Voxel Density Alignment (VDA) loss to enforce cross-view consistency. Our VDA loss regards the point cloud as an integral spatial distribution. It only enforces the numbers of points inside corresponding voxels of adjacent frames (voxels in the same color in Figure \ref{fig:1} (b)) to be consistent and does not penalize small spatial perturbation since the point still stays in the same voxel.

These two cross-view consistency losses exploit the temporal coherence in depth feature space and 3D voxel space for SS-MDE, both shifting the prior ``point-to-point'' alignment paradigm to the ``region-to-region'' one. Our method can achieve superior results than the state-of-the-art (SOTA). We conduct ablation experiments to demonstrate the effectiveness and robustness of the proposed losses.

\section{Related Work}
SS-MDE paradigm has become very popular in the community, which mainly takes advantage of cross-view consistency in monocular videos. In this section, we explore different categories of cross-view consistency used in previous self-unsupervised monocular depth estimation works.

\subsection{Photometric Cross-view Consistency}
The photometric cross-view consistency can be traced back to the Direct Method in SLAM (Simultaneous Localization and Mapping) optimizing camera poses through minimizing reprojection error, which skips the feature point extraction step in the traditional method and only depends on the difference in pixel intensity.
SFM-learner \cite{zhou2017unsupervised} is one of the first attempts to propose a self-supervised end-to-end network for training with monocular videos, which can jointly predict the depth and pose between consecutive frames. The core technique is using a spatial transformer network \cite{jaderberg2015spatial} to synthesize reference frames from source frames, which converts the depth estimation problem to a reprojection photometric error minimizing process. Geonet \cite{yin2018geonet} designs a joint learning framework of monocular depth, optical flow, and ego-motion estimation, which combines flow consistency with photometric consistency to model cross-view consistency. DF-Net \cite{zou2018df} also leverage the pixel-level consistency among multiple tasks including depth, optical flow, and motion segmentation estimation.  Gordon et al. \cite{Gordon_2019_ICCV} improve the photometric loss by simultaneously predicting a translation field and an occlusion-aware mask to exclude object motion and occlusion regions, respectively. But the occlusion-aware loss is calculated by comparing predicted depth values in consecutive frames, which is easily affected by the inaccuracy of estimated depth. \re{SGDdepth \cite{klingner2020self} introduces semantic guidance to solving photometric consistency violations caused by dynamic objects, via jointly learning depth estimation with supervised semantic segmentation task.} Monodepth2 \cite{godard2019digging} also proposes several schemes to improve the effectiveness of photometric loss, including an auto-masking loss and a minimum reprojection loss, yielding more accurate results. HRANet \cite{wang20223d} proposes intermediate view synthesis and pose augmentation to strengthen the restriction ability of photometric loss.
However, the photometric loss is not robust enough in some cases with violation factors and we believe areas with lower photometric loss can not necessarily guarantee more accurate depth and pose because of the ambiguity and low discriminability of photometric consistency loss, which is in line with the motivation of FeatDepth \cite{shu2020featdepth}. Moreover, the pixel-level photometric consistency may become invalid in challenging cases like moving objects region and the resulting occlusion area.

Following the most commonly used photometric consistency-based pipeline, some recent methods (DIFFNet \cite{zhou_diffnet}, MonoFormer \cite{bae2022deep}, MonoViT \cite{monovit} \blue{and SRD \cite{liu2023self})} explores to utilize the stronger network architectures including HRNet \cite{sun2019deep} and Vision Transformer (ViT) \cite{dosovitskiy2020image}, which further boosts the performance.

\subsection{Feature-level Cross-view Consistency}
Therefore, some researchers start to explore cross-view consistency at the feature level. Depth-VO-Feat \cite{zhan2018unsupervised} is a pioneering work to explore the combination of photometric consistency and feature-level consistency to generate temporally consistent depth estimation, taking binocular videos as input. Kumar et al. \cite{cs2018monocular} firstly combine Generative Adversarial Networks (GANs \cite{goodfellow2014generative,chen2020puppeteergan,chen2022d2animator}) with the self-supervised depth estimation architecture and use a discriminator to distinguish the synthetic reference frames and the real ones, which can also be regarded as a feature-level constraint. Following it, several more works are proposed to improve the GAN-based feature-level consistency loss \cite{zhao2020masked}. 
Except for utilizing deep representation learned in the depth estimation task separately, imposing semantic-aware features to enhance or align depth feature representation is a promising direction. Recent works \cite{li2021learning,Jung_2021_ICCV} propose to incorporate the semantic segmentation task to impose both feature-level implicit guidance and pixel-level explicit constraints.
Besides, FeatDepth \cite{shu2020featdepth} learns specific feature representations by a separate auto-encoder network in order to constrain cross-view consistency in feature space. These attempts prove the effectiveness of utilizing feature-level representation for cross-view consistency. We also explore feature-level consistency in the depth feature space but leverage the feature offsets estimated from temporal image frames to regularize the temporal coherency of estimated depth maps via Depth Feature Alignment (DFA) loss.

\subsection{3D Space Cross-view Consistency}
Besides the exploration of cross-view consistency in feature space, many works introduce additional 3D information to constrain geometric consistency. LEGO \cite{yang2018lego} presents a self-supervised framework to jointly estimate depth, normal, and edge, and uses normal as an additional 3D constraint to strengthen cross-view consistency constraint. Luo et al. \cite{luo2020consistent} leverage optical flow to find the corresponding 3D points in other frames and then build long-term 3D geometric constraints.
Similarly, GLNet \cite{chen2019self} simultaneously predicts depth and optical flow, and utilizes the predicted flow to build 3D points coupling to construct 3D points consistency and epipolar constraint. These two methods naturally build cross-view consistency in 3D space by imposing flow information among continuous frames, however, the accuracy heavily relies on the flow estimation which is an unsolved problem itself in different scenes, especially for those including occlusion and moving objects. 
Previous self-supervised depth estimation works started to impose additional information (i.e., normal and optical flow \cite{luo2020consistent}) to help enhance geometric restriction but hardly used 3D representation as to the 3D space cross-view consistency.
Vid2depth \cite{mahjourian2018unsupervised} constrains temporal cross-view consistency via 3D point cloud alignment based on the differentiable ICP method, which imposes 3D geometry information in the learning pipeline. However, point cloud alignment is a rigorous constraint to align corresponding 3D points and this loss is very sensitive to the points positions, which is fragile in scenes with moving objects and the resulting occlusion regions. By contrast, we propose Voxel Density Alignment (VDA) loss to impose 3D geometric information, which is robust and tolerant to the above challenging cases.

\section{METHODS}
\subsection{Preliminary}
\begin{figure*}
\begin{center}
\begin{minipage}[t]{\textwidth}
\centering
\includegraphics[width=16cm]{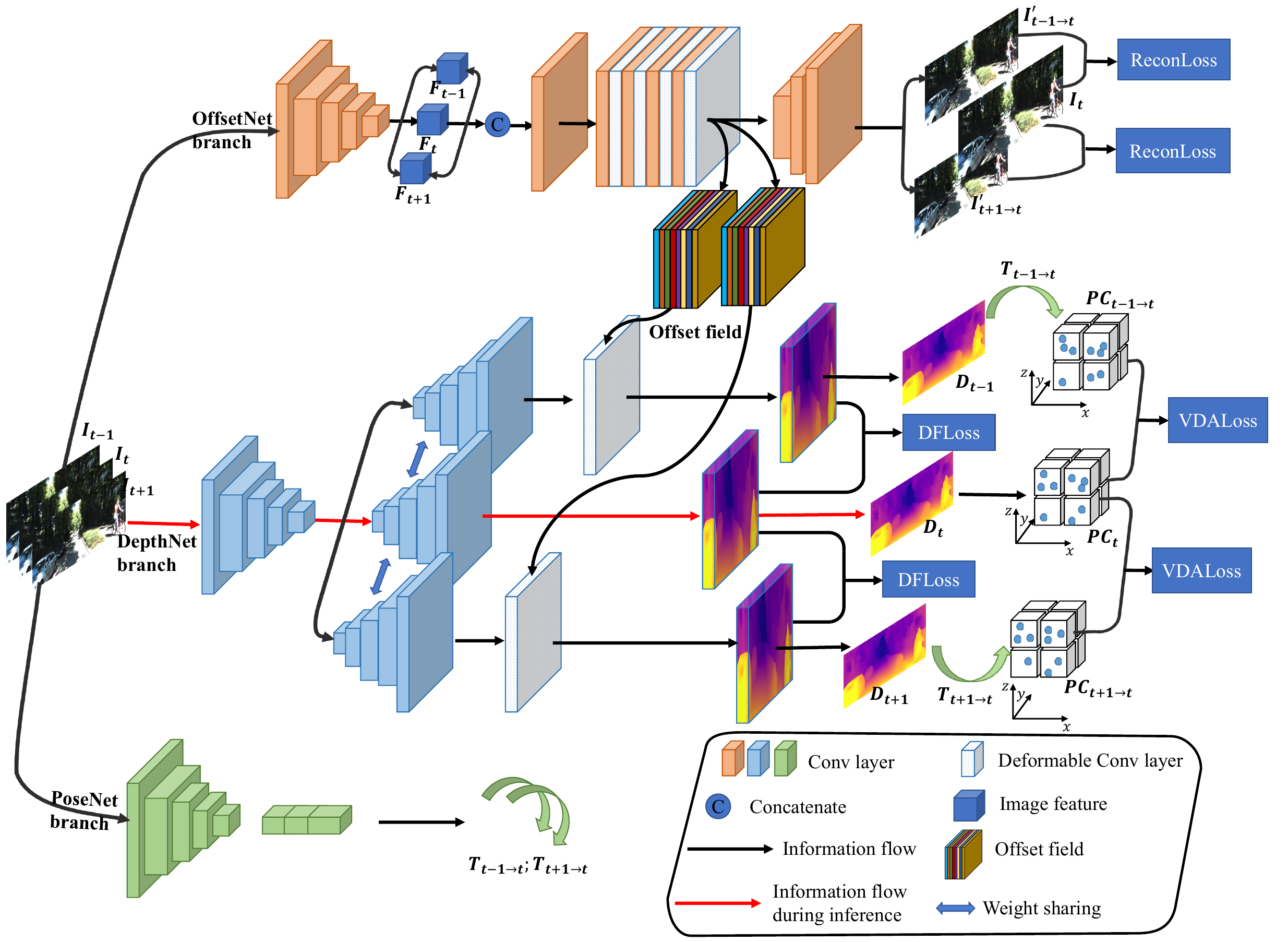}
\end{minipage}
\end{center}
\caption{An illustration of our learning framework, which consists of DepthNet, PoseNet, and OffsetNet for depth estimation, pose estimation, and alignment offset learning respectively. OffsetNet learns feature alignment offset field using self-supervised loss calculated by reconstructing reference from adjacent views with deformable convolutions. The learned offset field is then used to align temporal depth features learned from DepthNet. \re{The three branches in the framework are jointly optimized during training while only DepthNet is used during inference.}}
\label{fig:network}
\end{figure*}
\paragraph{Camera model}
The process of a camera mapping a point in 3D space to the 2D image plane can be described by a geometric model, basically the pinhole camera model. The mapping of a 3D point $P=(X, Y, Z)$ and its corresponding 2D point $p=(u,v)$ can be described as:
\begin{equation}
D(p)
\begin{bmatrix} u \\ v\\1\end{bmatrix}
= \begin{bmatrix} K\big|\textbf{0}\end{bmatrix}\begin{bmatrix} X \\ Y\\Z\\1\end{bmatrix}, 
\\
~{\rm where} ~ K=\begin{bmatrix} f_x &0&u_0\\ 0&f_y&v_0\\0&0&1\end{bmatrix}.
\label{eq:cameraProj}
\end{equation}
Matrix $K$ is the camera intrinsic matrix. $D(p)$ is the depth value at point $p$, i.e., the learning target of depth estimation task. Once the point $p$ and its depth value $D(p)$ are known, we can backproject it to get the corresponding 3D point $P$:
\begin{equation}
    P = D(p)K^{-1}p.
\end{equation}{}

\paragraph{2D cross-view consistency}
The essence of SS-MDE is using cross-view (in consecutive or stereo frames) consistency as the self-supervision signal. The most commonly used one is the photometric consistency, i.e.\re{,} assuming the intensity of 3D point $P$ projected in $I_t$ and $I_{t+m}$ is invariant:
\begin{equation}
    I_t(p_t)=I_{t+m}(p_{t+m}).
\end{equation}{}The projection point $p_{t+m}$ of $P$ in frame $I_{t+m}$ can be calculated from $p_t$ in frame $I_{t}$ and its depth $D(p_t)$, with the estimated transformation $T_{t\rightarrow t+m}$, by a differentiable warping function $\omega$: 
\begin{equation}
\begin{aligned}
p_{t+m}\sim\omega\left( KT_{t\rightarrow t+m}D(p_t)K^{-1}p_t\right).
\end{aligned}
\end{equation}{}Thus, the frame $\hat{I}_{t+m\rightarrow t}$ can be reconstructed from frame $I_{t+m}$:
\begin{equation}
    \hat{I}_{t+m\rightarrow t}(p) = I_{t+m}(p_{t+m}).
\end{equation}The photometric error minimization process is used to optimize depth and ego-motion estimation:
\begin{equation}
   L_{ph} =\sum_{p\in I_t}\left|I_{t}(p)-\hat{I}_{t+m\rightarrow t}(p)\right|.
\end{equation}{} \re{The photometric error adopted in previous works is usually the weighted sum of L1 and SSIM difference.}
To overcome the depth discontinuity, a smoothing term is often incorporated to add a regularization on depth maps in many previous works \cite{godard2017unsupervised,godard2019digging}:
\begin{equation}
    L_{sm} = \left|\partial_x\mu_{D_t}\right|e^{-|\partial_xI_t|}+\left|\partial_y\mu_{D_t}\right|e^{-|\partial_yI_t|},
\end{equation}
where $\mu_{D_t}$ is the inverse depth normalized by mean depth. \re{ $\partial_x\mu_{D_t}$ and $\partial_x\mu_{D_t}$ denote the disparity gradient among two directions.}

Although the photometric consistency effectively models the depth estimation task as a self-supervised problem, the photometric metric is not stable and robust based on the intensity invariance hypothesis, especially in complex outdoor scenes. Cases like moving objects, occlusion, and texture-less regions will mislead the optimization. Therefore, we propose two new cross-view consistency supervision from the perspective of deep feature space and 3D space.

\subsection{Depth feature alignment loss}
\begin{figure}
\begin{center}
\begin{minipage}[t]{0.45\textwidth}
\centering
\includegraphics[width=7cm]{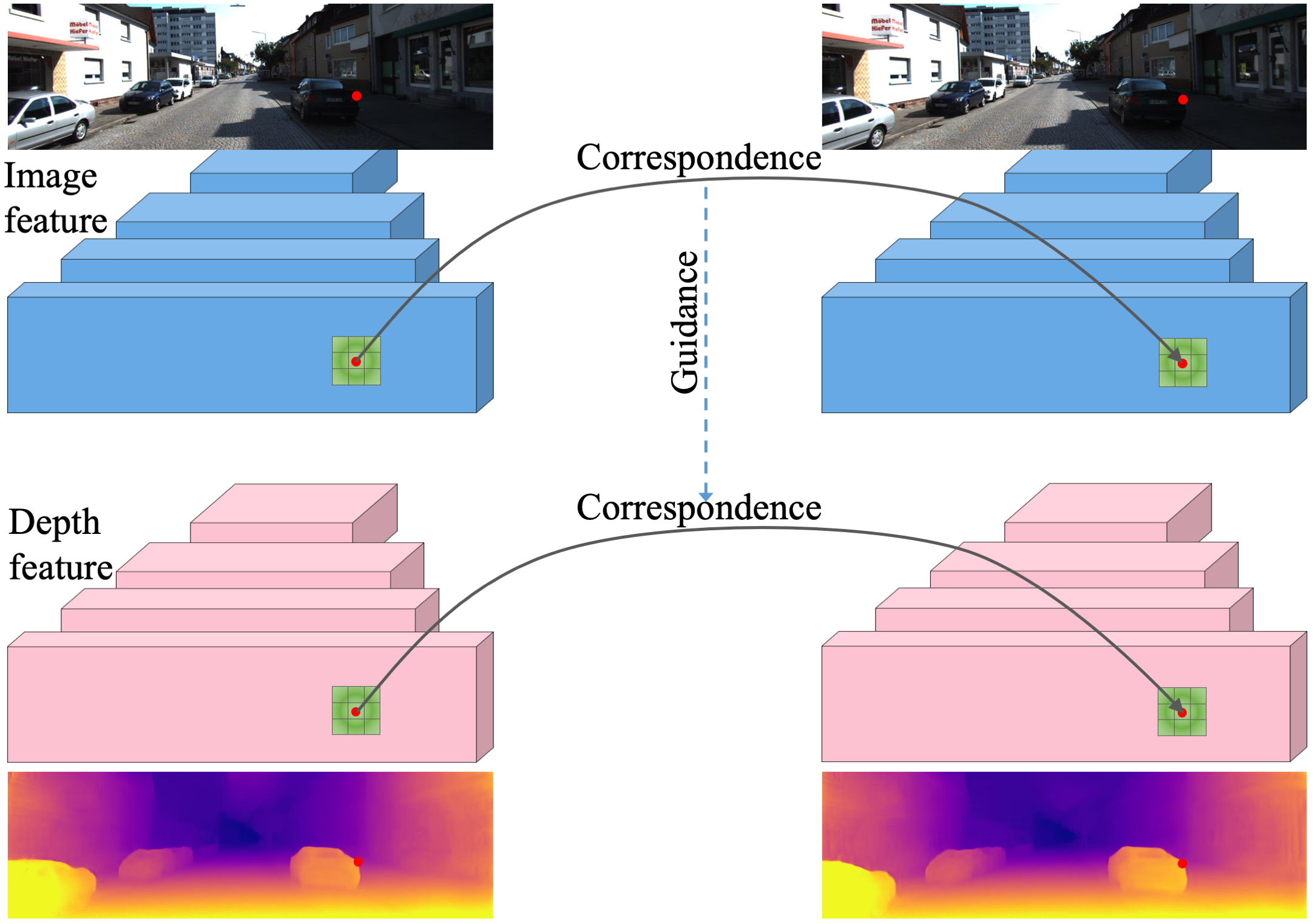}
\end{minipage}
\end{center}
\caption{Illustration of the guidance from the correspondence in RGB images to the correspondence in depth.}
\label{fig:correspondence}
\end{figure}

The motivation of DFA loss is that the coherence of consecutive depth frames is the same as the coherence of consecutive RGB frames, since the 2D pixel movement in either RGB or depth frames corresponds to the movement of the same 3D scene point, following the same projection described by Eq.~\eqref{eq:cameraProj}. As shown in the Figure \ref{fig:correspondence}, the correspondence learned from the RGB image can guide the maintenance of this correspondence during depth estimation. For example, a 3D point (red dot) is located on the edge of a black car. It will have many unique properties, such as an obvious depth change will occur around it. Then, in the next frame, it is still an edge point and still has this property. It is very natural to use optical flow to model the inter-frame movement. However, as discussed above, 2D photometric information is ambiguous and unreliable, which may also degrade optical flow estimation. Therefore, we propose to learn cross-view consistency from feature representation of RGB frames to guide the depth learning among corresponding frames. 

Different from \cite{shu2020featdepth} using a separate network to learn feature representation from RGB frames and aligning consecutive frames via differentiable feature warping, we learn the temporal feature alignment by reconstructing reference frame using its adjacent frames via deformable convolution networks \cite{dai2017deformable,tian2020tdan} in a totally self-supervised manner.

Given consecutive frames $I_t$ and $I_{t+m}$, the feature representations $F_t$ and $F_{t+m}$ are first learned via a feature extractor $\Phi$: $F_i = \Phi(I_i)$. The features extracted from adjacent frames are then taken as inputs into the deformable alignment network to learn the alignment offset $\Theta_{t+m\rightarrow t}$ between $F_t$ \re{and $F_{t+m}$ and obtain the aligned feature $\hat{F}_{t+m\rightarrow t}$}:
\begin{equation}
\begin{aligned}
\hat{F}_{t+m\rightarrow t},\Theta_{t+m\rightarrow t} = f_{\re{align}}(F_{t+m},F_t),
\end{aligned}
\end{equation}
\re{where $f_{align}$ denotes deformable alignment network, which consists of regular convolutions and deformable convolutions. The convolutional layer in deformable convolution is responsible for learning the 2D offsets and output the deformed feature map using bilinear interpolation. For each position $p$ on the aligned feature map $\hat{F}_{t+m\rightarrow t}$, it is calculated as:}
\begin{equation}
\begin{aligned}
\hat{F}_{t+m\rightarrow t}(p) = \sum_{k\in \Omega} \gamma(p_k)F_{t+m}(p+p_k+\Delta p_k).
\end{aligned}
\end{equation} 
\re{$\Theta_{t+m\rightarrow t} = \{\Delta p_k|k=1,...,|\Omega|\}$ denotes the offset learned by the deformable convolution, $\Omega$ is the kernel size and $p+p_k+\Delta$ is the $k$th learned additional offset at location $p+p_k$ by $\Delta p_k$. $p_k$ is the k-th sampling offset of a standard convolution with a kernel size of $n \times n$. For example, when $n=3$, we have $p_k \in \{(-1,-1),(-1,0),...,(1,1)\}$. The overall offset field learned by the deformable alignment network is a vector with $G\times 2N$ dimension for each pair of input images. $G$ is the deformable group number, which is set to 8 in our work. $2N$ represents the channel of each group offset field where the offset of each point is a two-dimensional vector, and it demonstrates the offset value on the x-direction and the y-direction respectively. $N$ means the square of kernel size $n\times n$.} 

\begin{figure}
\begin{center}
\begin{minipage}[t]{0.5\textwidth}
\centering
\includegraphics[width=8cm]{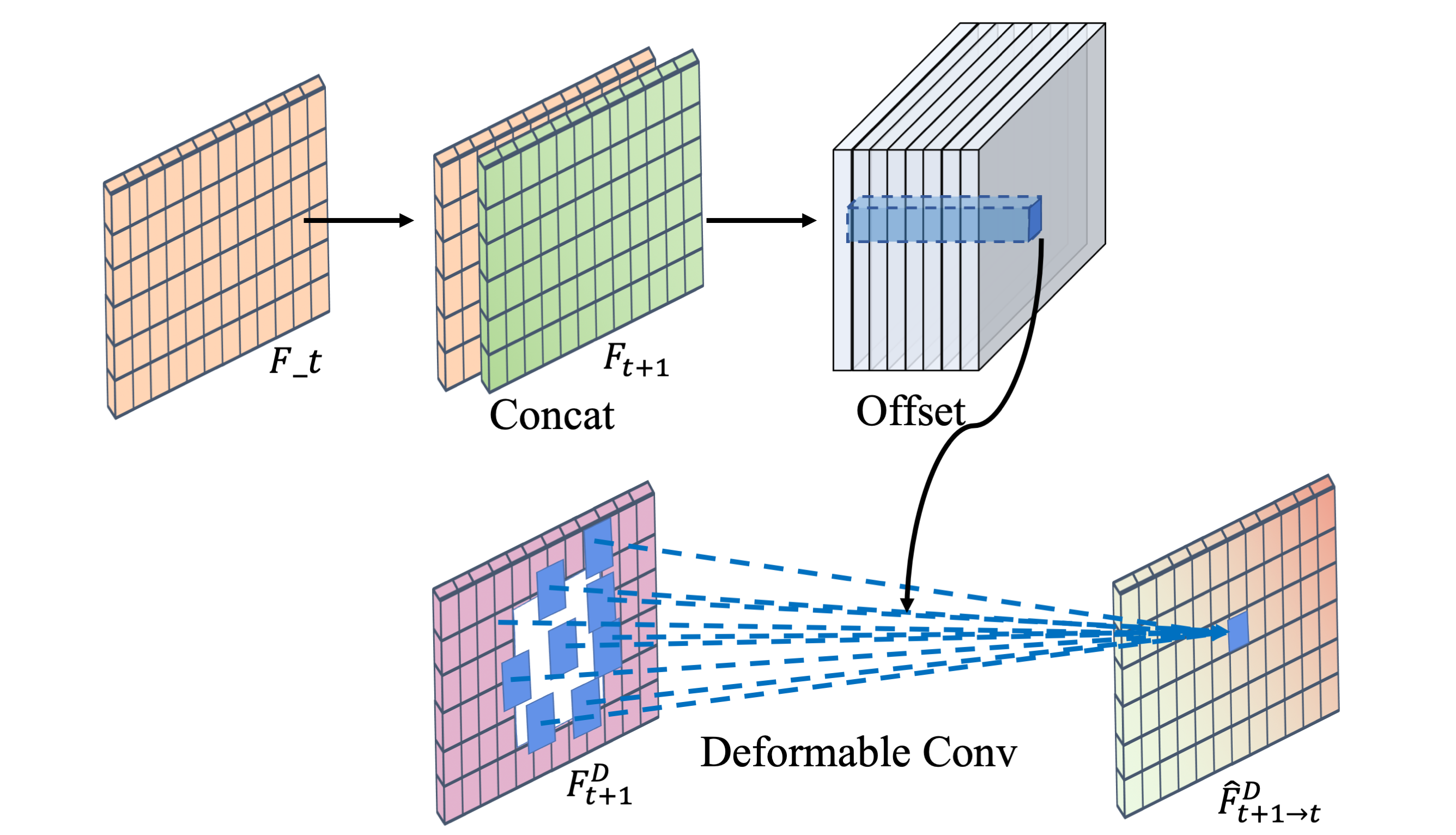}
\end{minipage}
\end{center}
\caption{Illustration of the key process of OffsetNet, which aims to learn feature alignment offsets from RGB frames. The learned offsets are then used to align depth features.}
\label{fig:offsetcore}
\end{figure}

In this way, the aligned feature $\hat{F}_{t+m\rightarrow t}$ is obtained, and the reference frame can be reconstructed from it:
\begin{equation}
    \hat I_{t+m\rightarrow t} = Re(\hat F_{t+m\rightarrow t}),
\end{equation}
where $Re$ denotes a reconstruction network simply consisting three \re{convolution layers}.
The feature alignment offset $\Theta_{t+m\rightarrow t}$ can be learned by minimizing the difference between the reconstructed and original reference frame, namely ReconLoss:
\begin{equation}
    L_{RE} = \left\|I_t - \hat I_{t+m\rightarrow t}\right\|^2.
\end{equation}
The learned offset is then used to conduct the temporal alignment of corresponding depth features:
\begin{equation}
    \hat F^D_{t+m} = f_{dc}\left(F^D_{t+m},\Theta_{t+m\rightarrow t}\right).
\end{equation}
Here, the deformable convolution $f_{dc}$ and $\Theta_{t+m\rightarrow t}$ are the same as the ones used in RGB features alignment to take advantage of the prior of feature alignment coherence.
The aligned depth feature $\hat F^D_{t+m}$ is enforced to be consistent with the estimated depth feature $F^D_t$ via DFLoss $L_{DF}$:
\begin{equation}
    L_{DF}=\left\|F^D_t-\hat F^D_{t+m}\right\|^2.
\end{equation}
Our DFA loss is a combination of the ReconLoss and DFLoss:
\begin{equation}
    L_{DFA}= L_{RE} + L_{DF}.
\end{equation}
The key process of DFA loss is shown in Figure \ref{fig:offsetcore}.
DFA loss governs consistent depth estimation using the temporal coherence learned from features instead of 2D photometric information. It is beneficial to overcome the vulnerability of photometric loss in cases like illumination variance because the feature-level alignment can model temporal alignment non-locally compared with the pixel-wise photometric alignment.
\red{\subsubsection{Analysis of DFA loss}\label{DFAlossAnalysis}
The cross-view consistency is usually enforced via the warping among adjacent frames, but DFA loss adopts an OffsetNet branch for learning the temporal alignment using the offset field from deformable convolutions. The previous excellent work FeatDepth \cite{shu2020featdepth} also utilizes the feature representation to improve the cross-view consistency. However, the extracted features of input images are just a kind of additional representation of photometric measurement. The cross-view alignment is still conducted by the warping among continuous frames, which highly relies on the predicted poses and makes it less robust to many challenging cases, e.g., low-texture and variant illumination regions. By contrast, our DFA loss adopts the OffsetNet branch to learn the cross-view alignment during training, and the learned feature-metric alignment is independent of the predicted poses, which makes our method more robust to the above challenging cases and can predict more temporally consistent and accurate depth.
}

\subsection{Voxel density alignment loss}
Due to the vulnerability of pixel-wise consistency supervision, Vid2depth \cite{mahjourian2018unsupervised} first impose 3D constraints by aligning two point clouds estimated in adjacent frames via Iterative Closest Point (ICP).
Enforcing the 3D geometry consistency of adjacent views seems to be reasonable and could be more effective compared with 2D consistency.
However, this point alignment constraint is too strict to be robust enough in challenging scenes with moving objects and occlusion.
We thus propose Voxel Density Alignment (VDA) loss as a new 3D cross-view consistency supervision that is robust to these challenging cases. Intuitively, the whole 3D space can be divided into the same number of voxels among consecutive views. Our VDA loss enforces the number of 3D points in corresponding voxels consistent between adjacent frames instead of forcing every corresponding point aligned. This means our VDA loss can be less affected by local object motion and occlusion.

To calculate the voxel density, we divide the 3D space into $N = N_x \times N_y \times N_z$ voxels\re{, using the Cartesian coordinate system shown in Figure \ref{fig:network}, with x- and y-axes as being horizontal and the z-axis as being verticald}. Point $P_i(x_i,y_i,z_i)\in \mathbb{R}^3$ in point cloud $PC= \lbrace{P_i}\rbrace^{n}_{i=1}$ will fall into voxel $V_j$, if $x_i\in[a_j,a_j+\Delta a), y_i\in[b_j,b_j+\Delta b),
z_i\in[c_j,c_j+\Delta c)$, where $(a_j, b_j, c_j, \Delta a, \Delta b, \Delta c)$ are a set of parameters presenting the spacial range of voxel $V_j$. Then, the voxel density can be calculated as:
\begin{equation}
C(V_j) = \sum^N_{i = 1}[P_i\in V_j],~~VD(V_j)= C(V_j)/n.
\end{equation}
Here, $[…]$ is the Iverson bracket. $[v]$ is defined to be $1$ if $v$ is true, and $0$ if it is false. $C(V_j)$ is a counting operation to obtain the number of points inside voxel $V_j$ and $n$ is the total number of 3D points. This is the naive implementation of VDA loss, which is easily understood but not differentiable due to the counting operation. We thus develop another technique to implement it in a differentiable and more efficient manner.

Specifically, once we estimated the depth map of a frame, the point cloud is easy to obtain. We first calculate a voxel index for each 3D point $P(x,y,z)$ according to the 3D position of the point:
\begin{equation}\label{Voxelization}
\begin{aligned}
\nu(P) = \left\lfloor\frac{x-x_{min}}{\Delta x}\right\rfloor+\left\lfloor\frac{y-y_{min}}{\Delta x}\right\rfloor N_x+\left\lfloor\frac{z-z_{min}}{\Delta z}\right\rfloor N_xN_y.
\end{aligned}
\end{equation}
Here, $N_x$, $N_y$, $N_z$ are the number of voxels along each axis, and $\Delta x = \frac{x_{max}-x_{min}}{N_x}$, $\Delta y = \frac{y_{max}-y_{min}}{N_y}$, $\Delta z = \frac{z_{max}-z_{min}}{N_z}$ are the shape parameters of voxels. Thus, point cloud $PC= \lbrace{P_i}\rbrace^{n}_{i=1}$ can be expressed as an $n$ dimension vector $V$. We then calculate the number of points in each voxel. We devise a function $g:\mathbb{R}^n \to \mathbb{R}^n$ to map $V$ to a counting 
vector $C= \lbrace{C_i}\rbrace^{n}_{i=1}$:
\begin{equation}
C_i = g_i(V)= n-\left\|sign(|V-i|)\right\|_1,
\end{equation}

In this way, the 3D point cloud can be represented as a voxel density vector: $\rho = C/n$. In conclusion, the calculation of the voxel density vector of frame $I_t$ from its estimated point cloud $PC_t$ is:
\begin{equation}
    \rho _t = \frac{1}{n} g(\nu(PC_t)).
\end{equation}
\re{$sign$ denote\re{s} the sign function,}
\re{\begin{equation}
 sign(x) =\left\{  
              \begin{array}{lr}  
              1, &  if\ x>0,\\  
              0, &  if\ x=0,\\  
              -1,&  if\ x<0 .   
              \end{array}  
 \right.  
 \end{equation}}
We refer to the Straight Through Estimator (STE) \cite{yin2018understanding} to differentiably implement the sign function \re{and get valid gradient during training}:
\re{
\begin{equation}  
 \left\{  
              \begin{array}{lr} 
              sign(r), &  fp\\  
              Htanh(r)=Clip(r,-1,1)=max(-1,min(1,r)),&  bp
              \end{array}  
 \right.  
 \end{equation}
here, $fp$ and $bp$ mean the forward pass and back-propagation process, and $r=(x-\frac{1}{2})\times 2$.}

The voxel density in one voxel can be regarded as the probability of 3D points situated in this 3D region. This representation of the 3D point cloud is more non-local and integral to express 3D geometry. To exploit the temporal coherence in voxel space, our VDA loss adopts KL divergence as the quantity to measure the voxel density vectors calculated from adjacent frames:
\begin{equation}
    L_{VD\re{A}}=D_{KL}\left(\rho_t||\rho_{t+m\rightarrow t}\right)=
    \sum _{i\in n}\rho_{t}(i) \log\left(\frac{\rho_t(i)}{\rho_{t+m\rightarrow t}(i)}\right).
\end{equation}
Here, $\rho_{t+m\rightarrow t} = \frac{1}{n}g(\nu(P_{t+m\rightarrow t}))$. $P_{t+m\rightarrow t}$ is the point cloud transformed from the frame $I_{t+m}$. By aligning the distribution of 3D point cloud using voxel density representation, we can shift the prior ``point-to-point'' alignment paradigm to the ``region-to-region'' one to constrain cross-view consistency, which will be more robust and tolerant to challenging cases like moving objects and occlusion.
\subsubsection{Analysis of VDA loss}
The effectiveness of our VDA loss mainly owes to the robust representation, i.e.\re{,} the voxel density. Here, we analyze the merit of the voxel density representation. Given two consecutive frames, the 3D geometry estimated from the two frames should be totally consistent after the ego-motion transformation if there is no inconsistent perturbation. However, in natural scenes, especially outdoor scenarios, violation cases such as moving people and vehicles are quite common.
Assuming we got $I_t$, $I_{t+\re{m}}$, their estimated depth map $D_t$, $D_{t+\re{m}}$, and their ego-motion transformation $T$, the most commonly used representation to depict the 3D geometry is the point cloud $P_t$ and $P_{t+m}$: 
\begin{equation}
P_t = D K^{-1}I_t,~~
P_{t+\re{m}} = D_{t+\re{m}}K^{-1}I_{t+\re{m}},~~
\hat P_t = T P_{t+\re{m}}.
\end{equation}
The prior point cloud loss measures the inconsistency of two estimated 3D point clouds via L1 norm:
\begin{equation}
    L_{pc} = \sum^n_{i=1}\left\|P_t(i) - \hat P_t(i)\right\|_1.
\end{equation}
L1 norm is sensitive to each element, which means this loss enforces each point to align with its corresponding point rigidly.
Sometime there is an object motion between two frames, which means the existing of a small perturbation on some points. Assuming point $P_1(x_1,y_1,z_1)$ moves $\delta_{x_1}, \delta_{y_1}, \delta_{z_1}$ in axis $x, y, z$ respectively, the cross-view consistency loss can be calculated via point cloud loss as:
\begin{equation}
    L_{pc} = |\delta_{x_1}| + |\delta_{y_1}| + |\delta_{z_1}|.
\end{equation}
Differently, our VDA loss pays more attention to the spatial positions of 3D points, measuring the inconsistency of corresponding groups of points. Grouping operation is realized via putting points into different voxels according to their 3D positions:
\begin{equation}
    L_{v} = \left\|\nu(P_t(i)\re{)}-\nu(\hat P_t(i))\right\|_1,
\end{equation}
where the ``voxelization'' process $\nu(P)$ is calculated as Eq.~\eqref{Voxelization}. When there is a small perturbation $\delta_{x_1}, \delta_{y_1}, \delta_{z_1}$ in $P_1$, $L_v$ can be calculated as:
\begin{equation}
\begin{aligned}
L_{v} &= \left\lfloor\frac{x+\delta x-x_{min}}{\Delta x}\right\rfloor+\left\lfloor\frac{y+\delta y-y_{min}}{\Delta y}\right\rfloor N_x\\&+\left\lfloor\frac{z+\delta z-z_{min}}{\Delta z}\right\rfloor N_xN_y\\
   & - \left(\left\lfloor\frac{x-x_{min}}{\Delta x}\right\rfloor+\left\lfloor\frac{y-y_{min}}{\Delta y}\right\rfloor N_x+\left\lfloor\frac{z-z_{min}}{\Delta z}\right\rfloor N_xN_y\right).
\end{aligned}
\end{equation}
Taking $z$ as for example, because there is a floor operation, only when $\frac{z+\delta z-z_{min}}{\Delta z}-\frac{z-z_{min}}{\Delta z} > 1$, the value of $L_{v}$ can be non-zero, which means $\delta z$ need to be larger than $\Delta z$. Therefore, the small perturbation will not change the voxel index $\nu(P_1)$, so that $L_v$ maintains $0$ and $L_{VD}$ maintains its stability. For an intuitive explanation, although the object has a small motion, this whole object still stays in the original voxel as shown in Figure \ref{fig:1}. Therefore, our VDA loss is more robust to object motions than the point cloud alignment loss.
\begin{table*}[t]
\small
\begin{center}
\caption{Quantitative performance of single depth estimation on KITTI eigen test set. For a fair comparison, all the results are evaluated at the maximum depth threshold of 80m. 
All methods are evaluated with raw LiDAR scan data. ``$\dagger$" means updated results after publication. 
Bold is the best indicator and underlines indicate the second-best results.
$\delta_1, \delta_2, \delta_3$ denote $\delta < 1.25, \delta < 1.25^2, \delta < 1.25^3$, respectively. \re{The column ``train'' means training manners, with ``M'' denonte self-supervised monocular training and ``M\&Seg'' denotes self-supervised monocular training together with supervised segmentation training.}}
\label{tab:1}
\setlength{\tabcolsep}{0.005\linewidth}
\begin{tabular}{ccccccccccc}
\toprule
\multirow{2}{*}{\!\!\!Methods\!\!\!}&\multirow{2}{*}{Train} &\multirow{2}{*}{Backbone}& \multirow{2}{*}{Resolution}&
 \multicolumn{4}{c}{Error metric$\downarrow$} & \multicolumn{3}{c}{Accuracy metric$\uparrow$} \\
\cmidrule(r){5-8} 
\cmidrule(r){9-11} 

  &&& &Abs Rel      & Sq Rel &   RMSE
&  RMSE log \!\!\!     &  $\delta_1$   &   $\delta_2 $
&  $\delta_3$      \\
\midrule
SFMlearner \cite{zhou2017unsupervised} (CVPR 2017)$\dagger$&M&DispNet &	$416\times128$	&0.183	&1.595	&6.709	&0.270	&0.734	&0.902	&0.959\\
Mahjourian et al. \cite{mahjourian2018unsupervised} (CVPR 2018) &M& DispNet &$416\times128$ &0.163	&1.240	&6.220	&0.250	&0.762	&0.916	&0.968\\
GeoNet \cite{yin2018geonet} (CVPR 2018)$\dagger$	&M&ResNet50 &$416\times128$	&0.149	&1.060	&5.567	&0.226	&0.796	&0.935	&0.975\\
DDVO \cite{wang2018learning}  (CVPR 2018)	&M&DispNet &	$416\times128$	&0.151	&1.257	&5.583	&0.228	&0.810	&0.936	&0.974\\
DF-Net \cite{zou2018df} (ECCV 2018)	&M& ResNet50 &$576\times160$		&0.150	&1.124	&5.507	&0.223	&0.806	&0.933	&0.973\\
Struct2depth \cite{casser2019depth}(AAAI 2019)	&M& DispNet  &$416\times128$		&0.141	&1.026	&5.142	&0.210	&0.845	&0.845	&0.948 \\
SC-SFMlearner \cite{bian2019unsupervised} (NeurIPs 2019)	&M&DispResNet \!\!\!& $832\times256$		&0.137	&1.089	&5.439	&0.217	&0.830	&0.942	&0.975\\
HR \cite{zhou2019unsupervised} (ICCV 2019)&M&	ResNet50& $1248\times384$		&0.121	&0.873	&4.945	&0.197	&0.853	&0.955	&0.982\\
Monodepth2 \cite{godard2019digging}  (ICCV 2019)	&M& ResNet18 & $640\times192$	&0.115	&0.903	&4.863	&0.193	&0.877	&0.959	&0.981\\
PackeNet \cite{Guizilini_2020_CVPR}  (CVPR 2020)&M& PackNet &$640\times 192$  &0.111	&0.785	&4.601	&0.189	&0.878	&0.960	&0.982\\

TrianFlow \cite{zhao2020towards} (CVPR 2020)&M&ResNet18&$832\times 256$&0.113&0.704&4.581&0.184&0.871&0.961&\underline{0.984}\\
Johonston et al. \cite{johnston2020self}  (CVPR 2020)&M &ResNet101&$640\times 192$&0.106 &0.861 &4.699 &0.185&0.889&0.962&0.982\\

FeatDepth \cite{shu2020featdepth}  (ECCV 2020)&M& ResNet50 &$1024\times320$ &0.104 &0.729 &4.481 &0.179 &0.893 &\underline{0.965} &\underline{0.984}\\
MLDA-Net \cite{song2021mlda} (TIP 2021) &M&ResNet50&$640\times 192$  &0.110	&0.824	&4.632	&0.187	&0.883	&0.961	&0.982\\

HR-Depth \cite{Lyu_Liu_Wang_Kong_Liu_Liu_Chen_Yuan_2021}  (AAAI 2021)&M&HRNet &$640\times 192$  &0.109	&0.792	&4.632	&0.185	&0.884	&0.962	&0.983\\
R-MSFM6 \cite{zhou2021r} (ICCV 2021)&M &ResNet18 &$640\times 192$		& 0.112	& 0.806	& 4.704	& 0.191	& 0.878	& 0.960	& 0.981\\
 Wang et al. \cite{Wang_2021_ICCV} (ICCV 2021)& M& ResNet18& $640\times 192$& 0.109& 0.779& 4.641& 0.186& 0.883& 0.962& 0.982\\
CADepth \cite{yan2021channel} (3DV 2021)& M& ResNet50& $640\times 192$& 0.105& 0.769& 4.535& 0.181& 0.892& 0.964& 0.983\\
DIFFNet \cite{zhou_diffnet} (BMVC 2021)& M& HRNet& $640\times 192$& 0.102& 0.749& 4.445& 0.179&\underline{0.897}&\underline{0.965}& 0.983\\
MonoFormer \cite{bae2022deep}(AAAI 2023) & M& ViT& $640\times 192$& 0.104& 0.846& 4.580& 0.183& 0.891& 0.962& 0.982\\
MonoViT \cite{monovit} (3DV 2022)& M & ViT& $640\times 192$&\textbf{ 0.099}&\underline{ 0.708}&\underline{4.372}&\underline{0.175}&\textbf{ 0.900}&\textbf{ 0.967}&\underline{0.984}\\
\blue{HRANet \cite{wang20223d} {(TCSVT2022)}}&\blue{M}& \blue{ResNet18}& \blue{$640\times 192$}& \blue{0.109}& \blue{0.790}& \blue{4.656}& \blue{0.185}& \blue{0.882}& \blue{0.962}& \blue{0.983}\\
\blue{SRD \cite{liu2023self} {(TCSVT2023)}}& \blue{M}&\blue{ViT }& \blue{$640\times 192$}		& \blue{\textbf{0.099}}	& \blue{\textbf{0.659}}	& \blue{\textbf{4.314}}	&\blue{\underline{ 0.174}}	& \blue{0.898} & \blue{\textbf{0.967}}	&\blue{\textbf{ 0.985}}\\
\textbf{Ours}	& {M}&ViT & $640\times 192$		& \textbf{0.099}	& \textbf{0.667}	& \textbf{4.338}	&\textbf{ 0.173}	& \textbf{0.900}	& \textbf{0.967}	&\textbf{ 0.985}\\
\bottomrule
Monodepth2\cite{godard2019digging}  {(ICCV 2019)}	& {MS}& ResNet18 & $640\times192$	&0.106	&0.818	&4.750	&0.196	&0.874	&0.957	&0.979\\
HR-Depth \cite{Lyu_Liu_Wang_Kong_Liu_Liu_Chen_Yuan_2021}   {(AAAI 2021)}& {MS}&HRNet &$640\times 192$  &0.107	&0.785	&4.612	&0.185	&0.887	&0.962	&0.983\\
CADepth \cite{yan2021channel} (3DV 2021)& {MS}& {ResNet50}& {$640\times 192$}& {0.102}& {0.752}& {4.504}& {0.181}& {0.894}& {0.964}& {0.983}\\
DIFFNet \cite{zhou_diffnet} (BMVC 2021)& {MS}& {HRNet}& {$640\times 192$}& {0.101}& {0.749}& {4.445}&\underline{0.179}&\underline{0.898}&\underline{0.965}& {0.983}\\
MonoViT \cite{monovit}  {(3DV 2022)}& {MS} & {ViT} &$640\times 192$ &\underline{0.098} &\underline{0.683} &\underline{4.333} & {0.174}& {\textbf{0.904}}&\textbf{ {0.967}}&\underline{0.984}\\
\textbf{Ours}& {MS}&ViT & $640\times 192$		& \textbf{0.096}	& \textbf{0.666}	& \textbf{4.328}	&\textbf{ 0.174}	& \textbf{0.904}	& \textbf{0.967}	&\textbf{ 0.985}\\
\bottomrule

\end{tabular}
\end{center}
\end{table*}
\subsection{Overall learning pipeline}
In this paper, our method adopts DFA loss and VDA loss as additional cross-view consistency to the widely used photometric loss and smooth loss. Therefore, the total loss is:
\begin{equation}
    L = L_{ph} +\alpha L_{sm} + \beta L_{DFA} +\gamma L_{VDA}.
\end{equation}
{Here, $\alpha, \beta, \gamma$ is $0.01,0.05,0.05$, respectively. The weights of the photometric loss and smooth loss are set according to our baseline method Monodepth2 \cite{godard2019digging}. And the parameters $\beta$ and $\gamma$ are used to keep the loss values in a similar scale as the basic photometric loss $L_{ph}$.} The parameters $N_x$,$N_y$,$N_z$ are 40, 24, and 40 in our work. Our implementation of photometric and smooth loss follows the baseline method \cite{godard2019digging}.

\begin{figure*}[htbp]
\begin{center}
\begin{minipage}{0.96\textwidth}

\includegraphics[ width=3.4cm,valign=t]{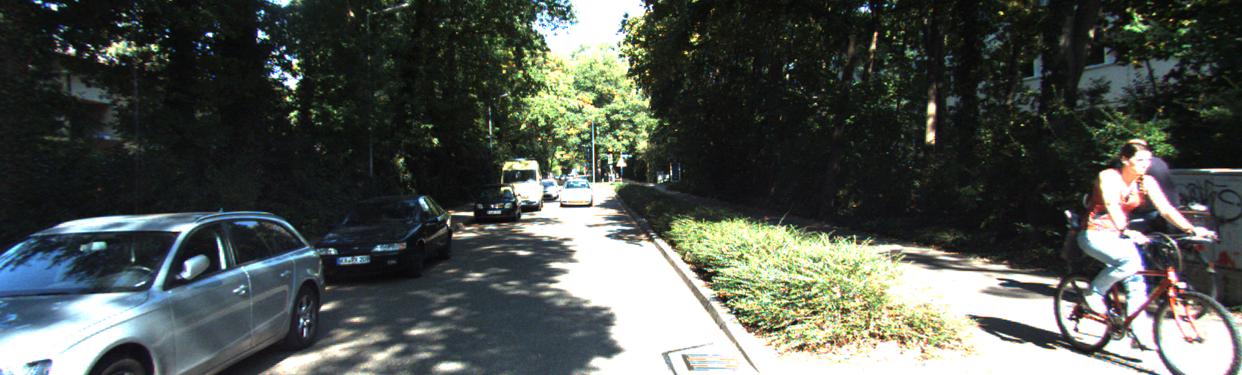}\includegraphics[ width=3.4cm,valign=t]{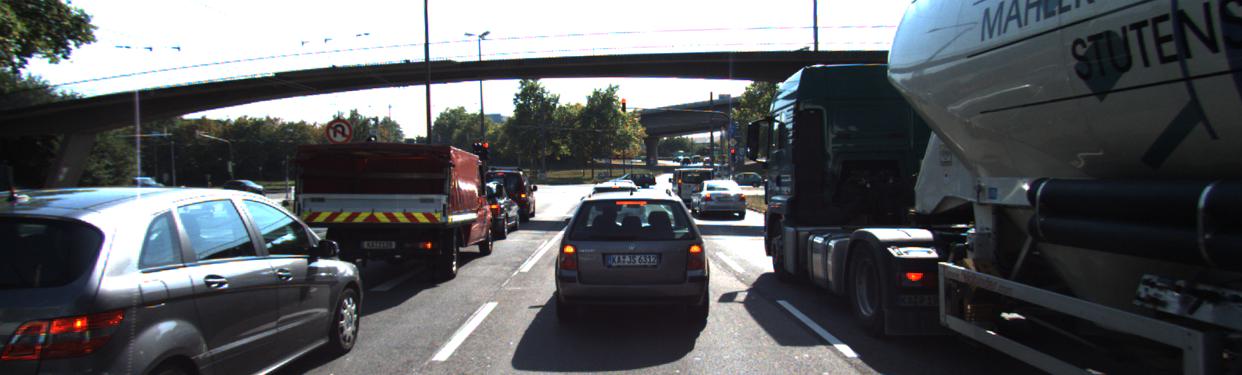}\includegraphics[ width=3.4cm,valign=t]{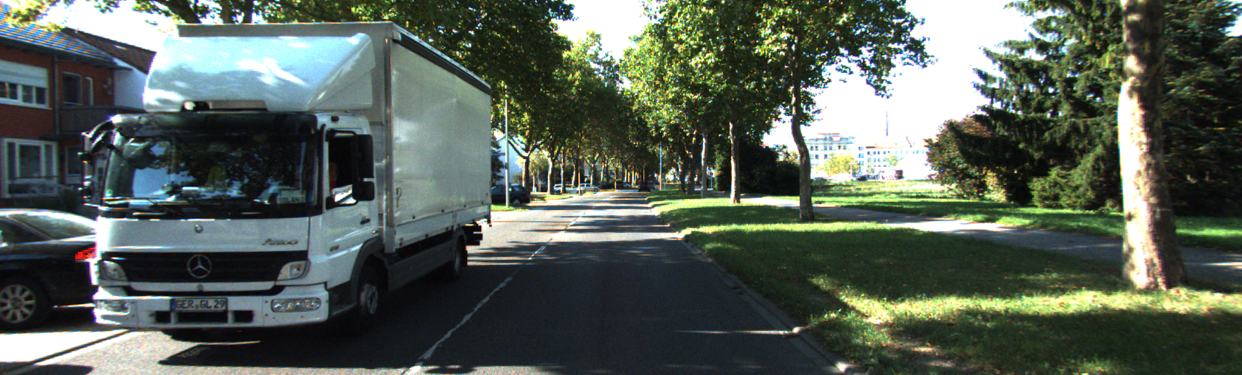}\includegraphics[ width=3.4cm,valign=t]{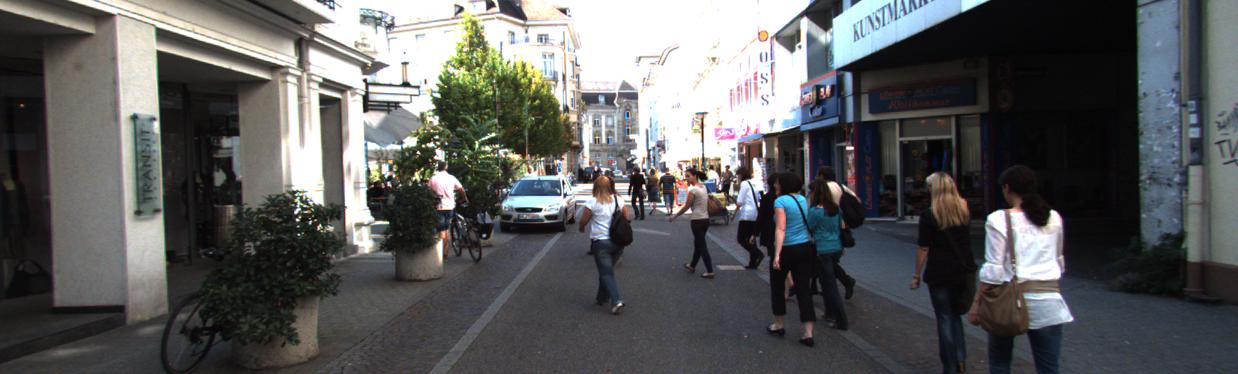}\includegraphics[ width=3.4cm,valign=t]{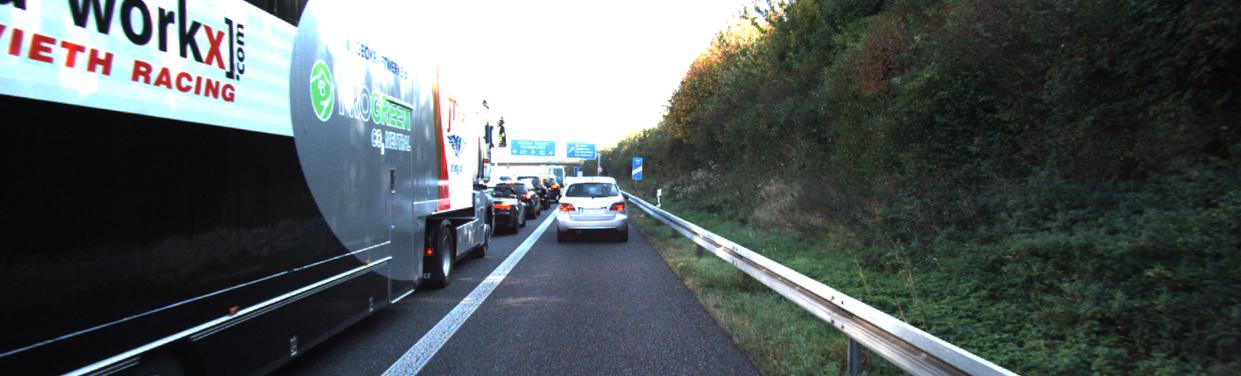}

\includegraphics[ width=3.4cm,valign=t]{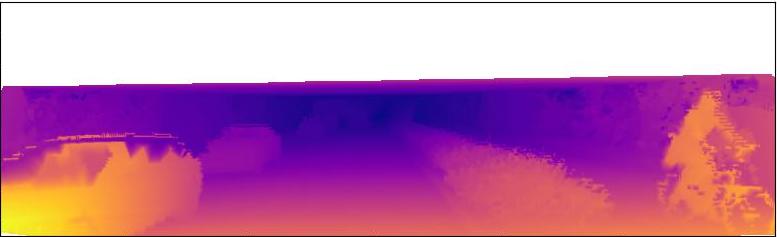}\includegraphics[ width=3.4cm,valign=t]{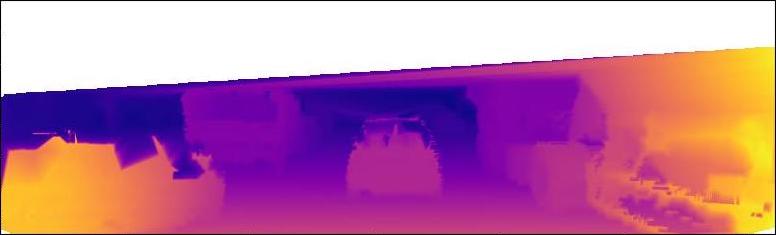}\includegraphics[ width=3.4cm,valign=t]{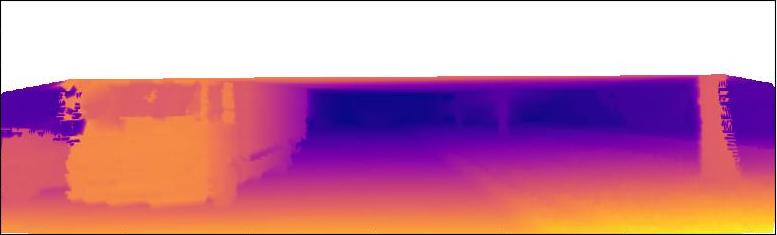}\includegraphics[ width=3.4cm,valign=t]{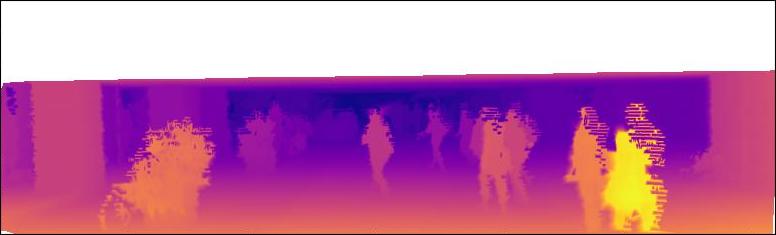}\includegraphics[ width=3.4cm,valign=t]{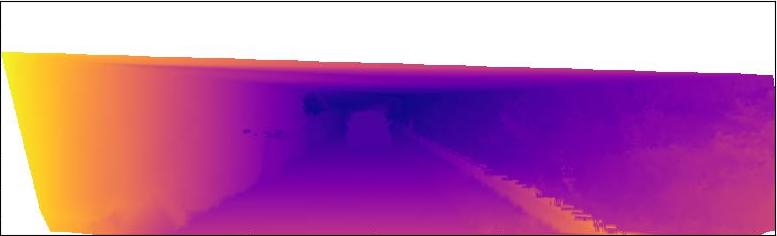}

\includegraphics[ width=3.4cm,valign=t]{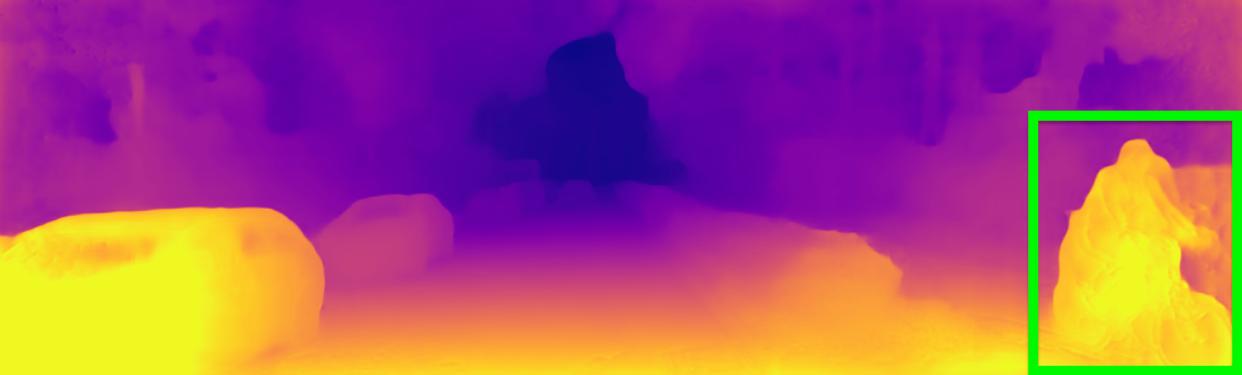}\includegraphics[  width=3.4cm,valign=t]{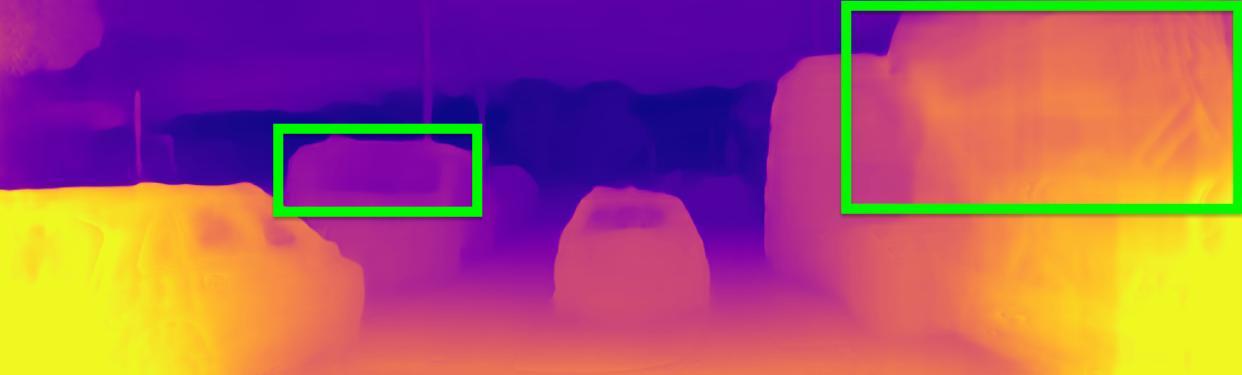}\includegraphics[  width=3.4cm,valign=t]{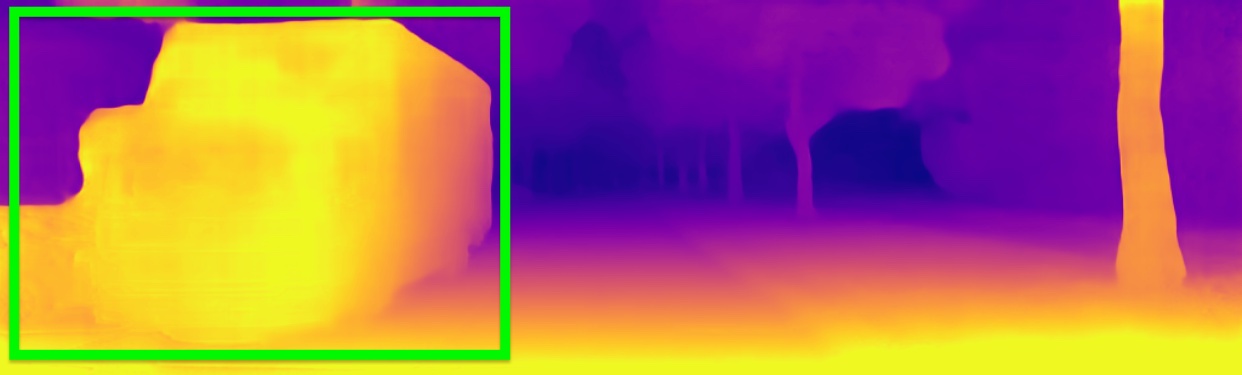}\includegraphics[  width=3.4cm,valign=t]{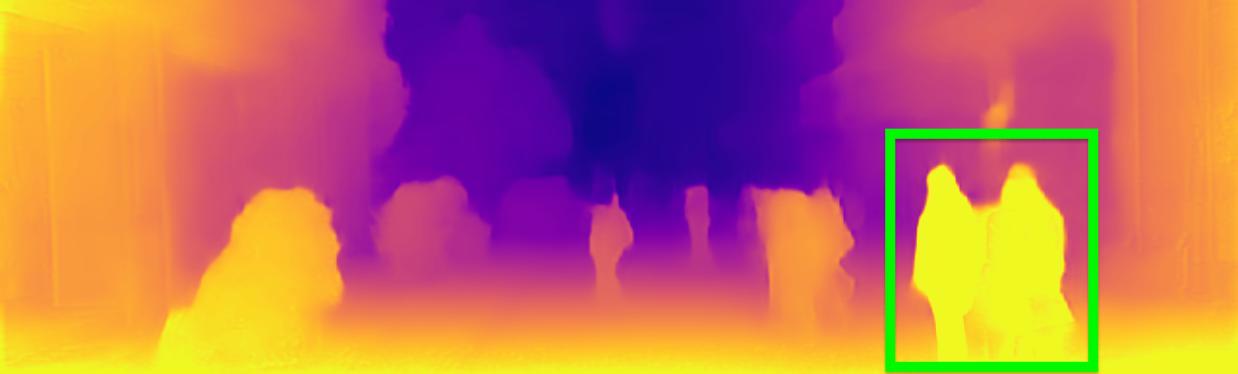}\includegraphics[  width=3.4cm,valign=t]{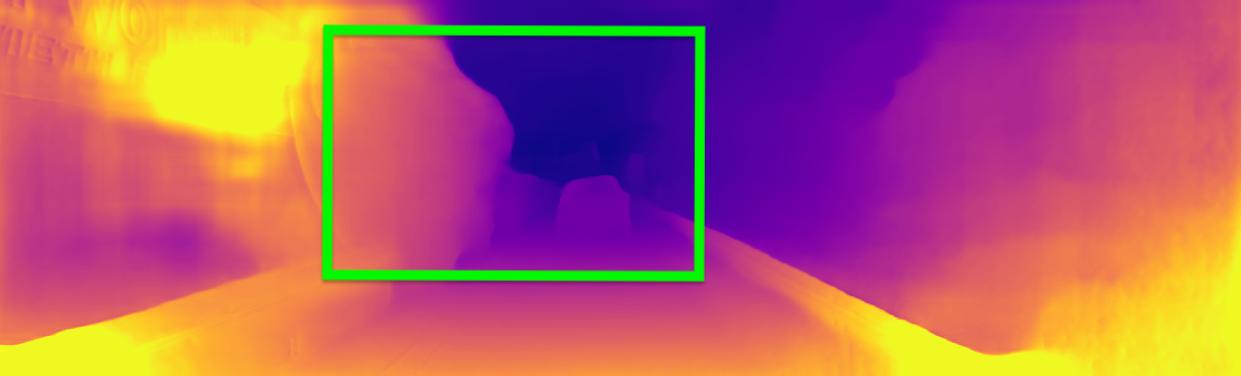}

\includegraphics[width=3.4cm,valign=t]{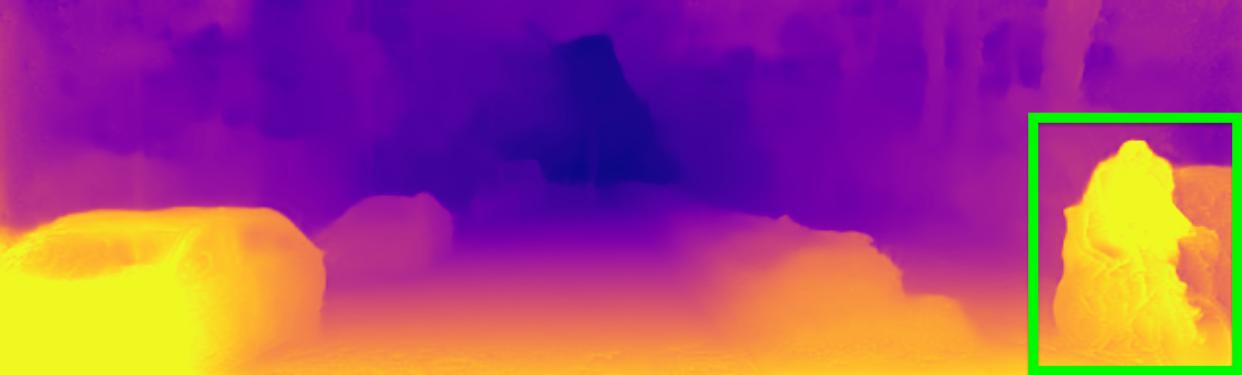}\includegraphics[  width=3.4cm,valign=t]{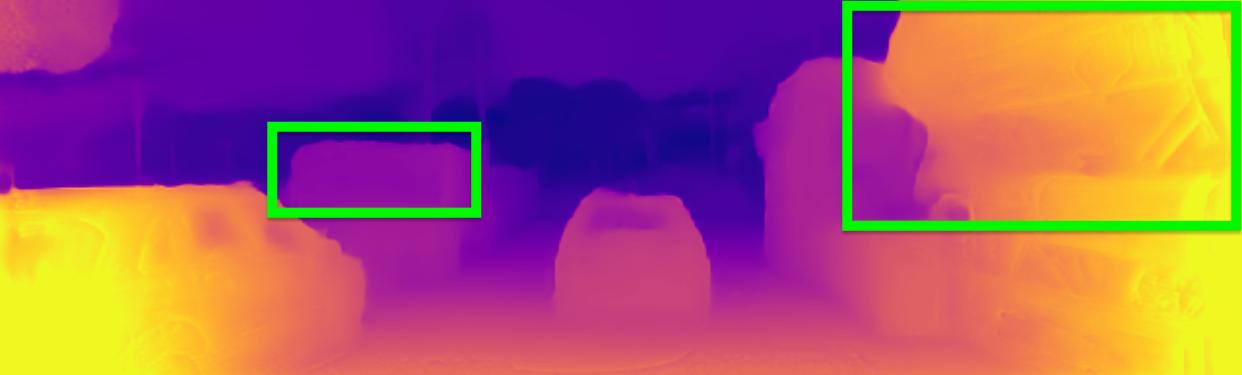}\includegraphics[width=3.4cm,valign=t]{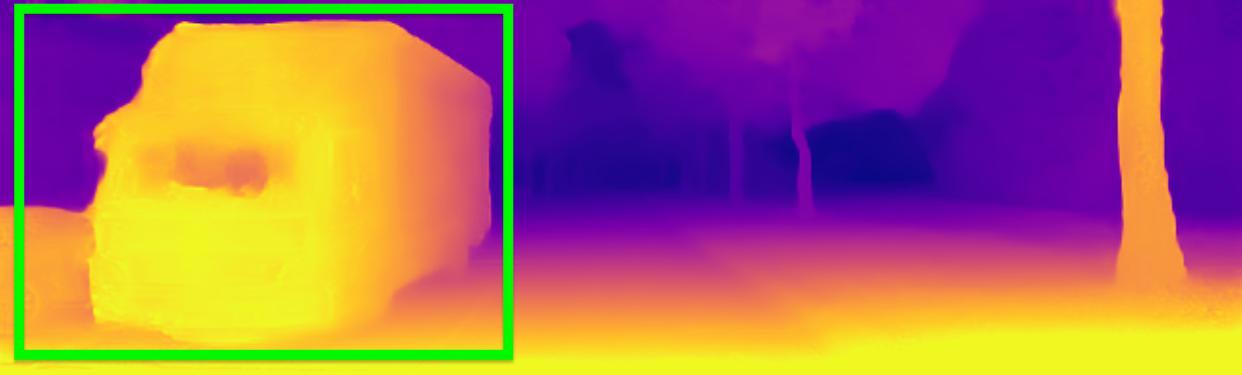}\includegraphics[width=3.4cm,valign=t]{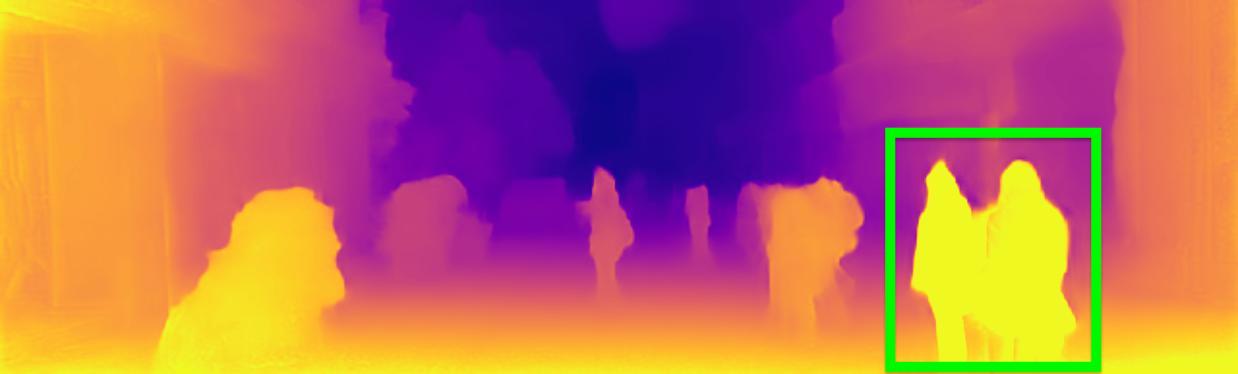}\includegraphics[width=3.4cm,valign=t]{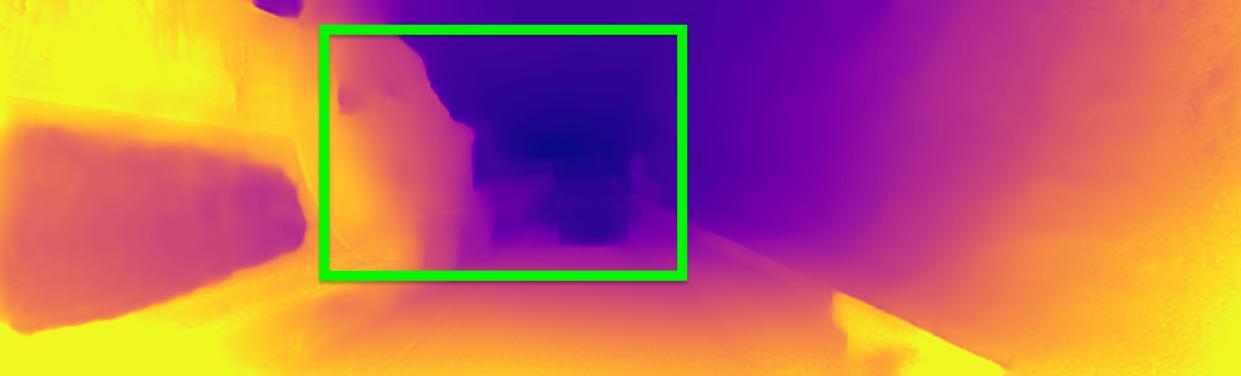}

\includegraphics[ width=3.4cm,valign=t]{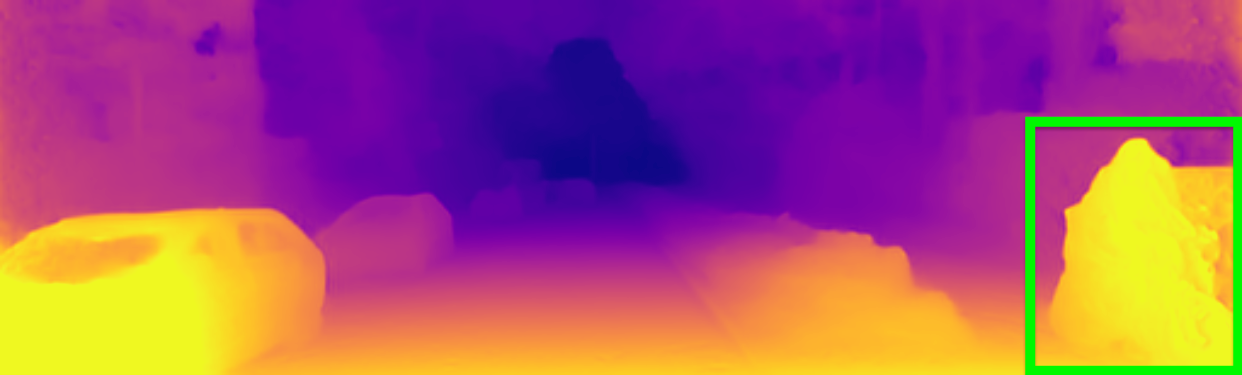}\includegraphics[  width=3.4cm,valign=t]{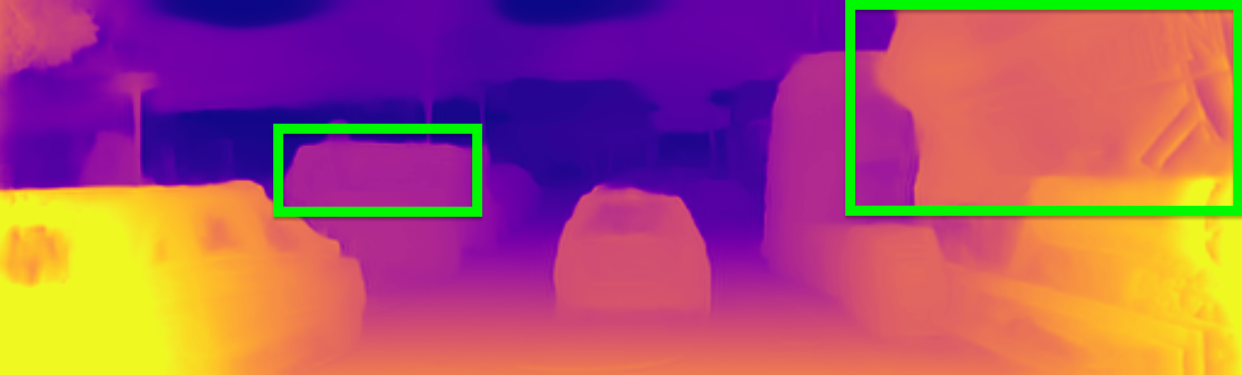}\includegraphics[  width=3.4cm,valign=t]{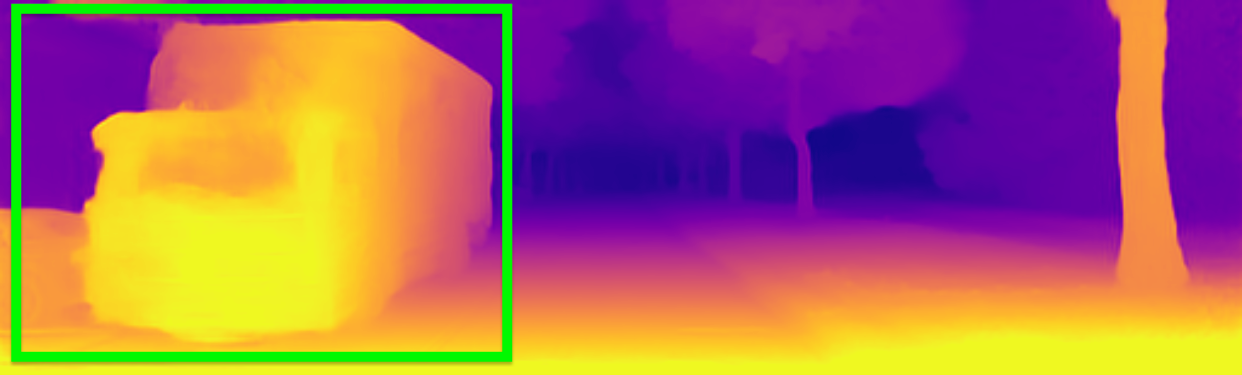}\includegraphics[  width=3.4cm,valign=t]{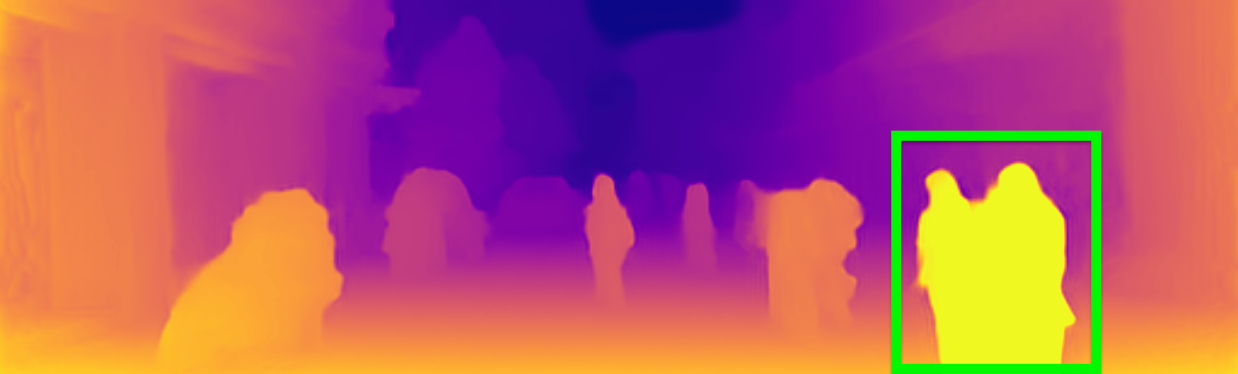}\includegraphics[  width=3.4cm,valign=t]{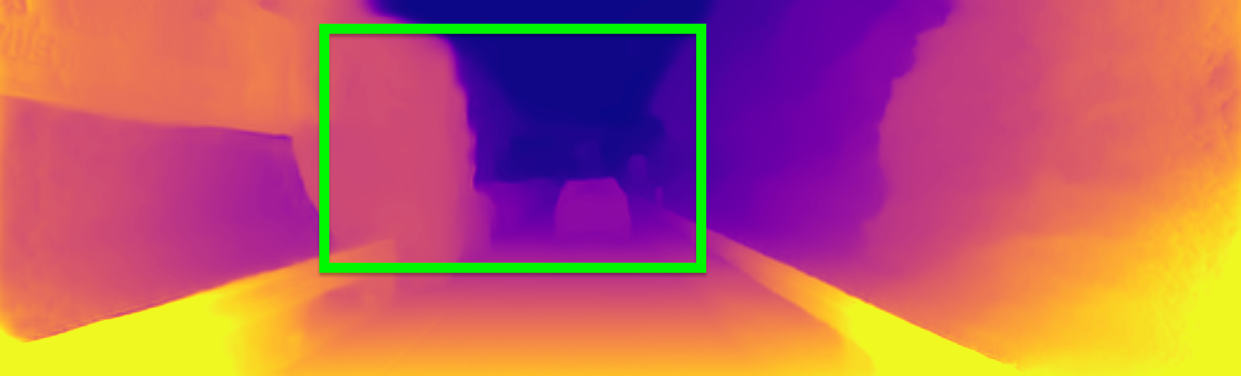}

\includegraphics[ width=3.4cm,valign=t]{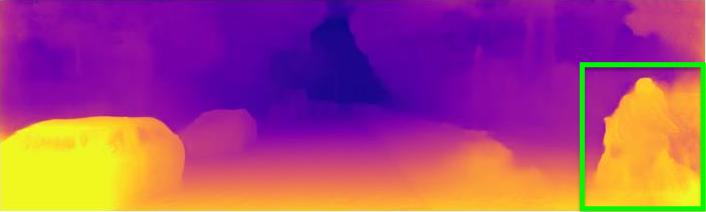}\includegraphics[  width=3.4cm,valign=t]{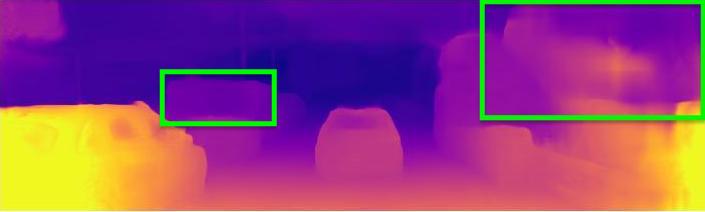}\includegraphics[  width=3.4cm,valign=t]{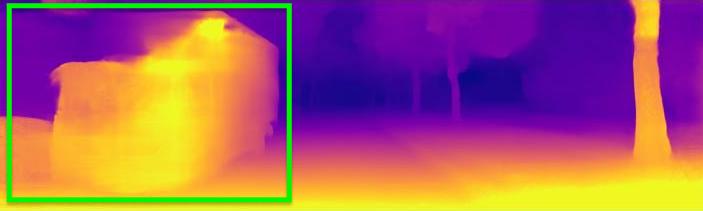}\includegraphics[  width=3.4cm,valign=t]{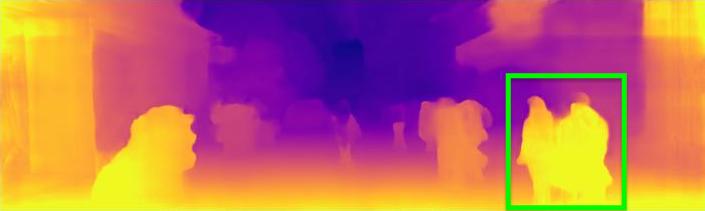}\includegraphics[  width=3.4cm,valign=t]{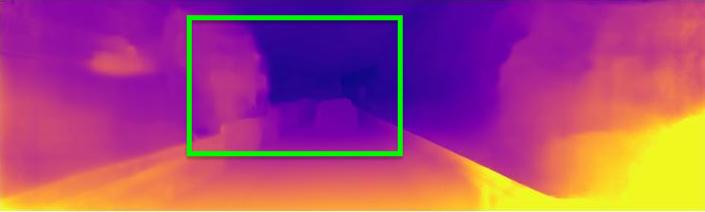}

\includegraphics[ width=3.4cm,valign=t]{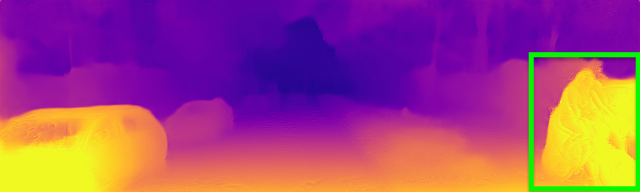}\includegraphics[  width=3.4cm,valign=t]{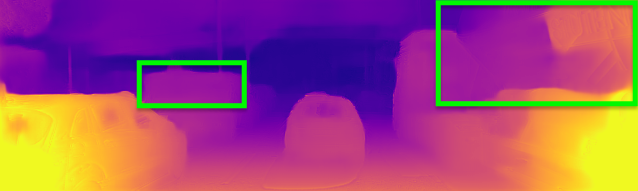}\includegraphics[  width=3.4cm,valign=t]{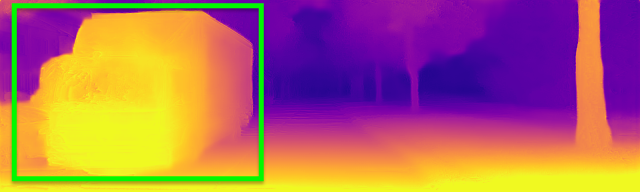}\includegraphics[  width=3.4cm,valign=t]{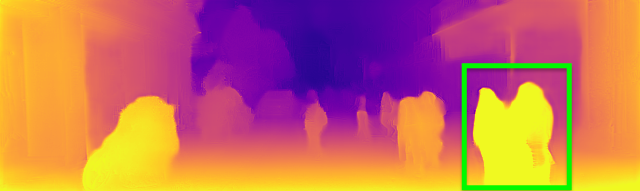}\includegraphics[  width=3.4cm,valign=t]{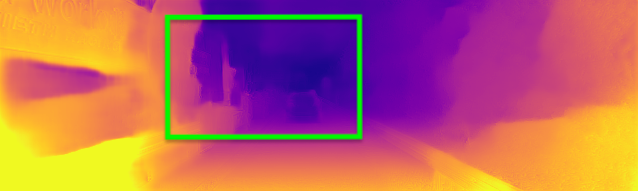}

\includegraphics[  width=3.4cm,valign=t]{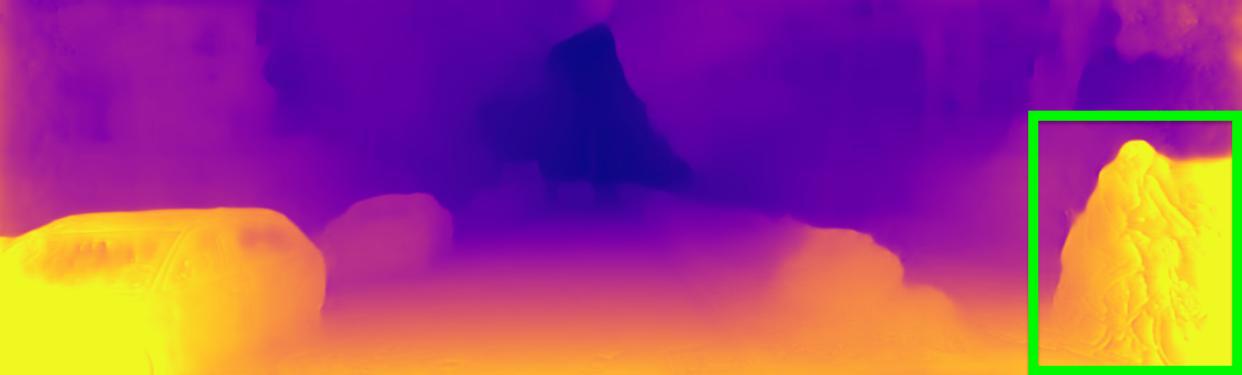}\includegraphics[  width=3.4cm,valign=t]{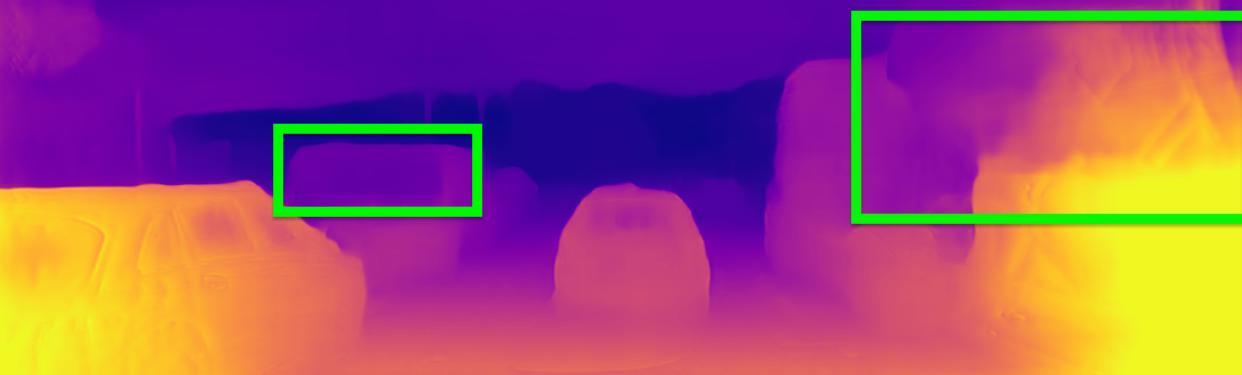}\includegraphics[  width=3.4cm,valign=t]{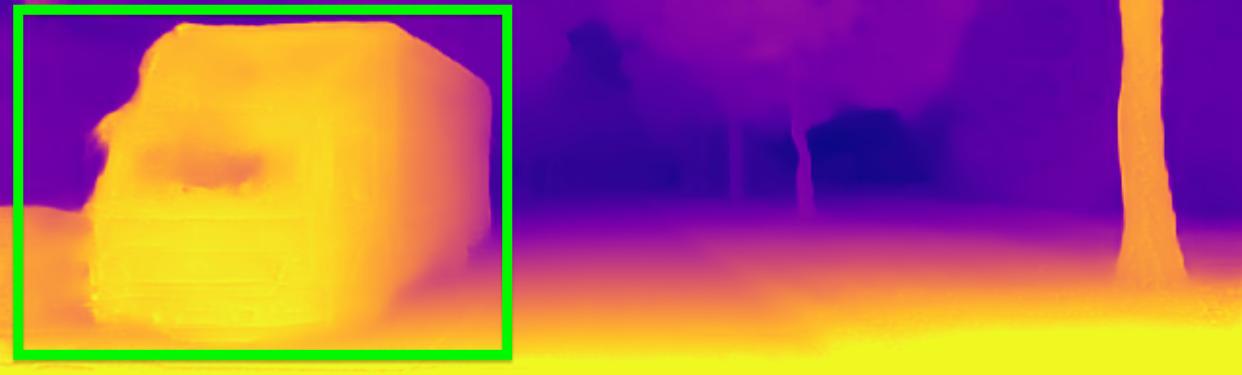}\includegraphics[  width=3.4cm,valign=t]{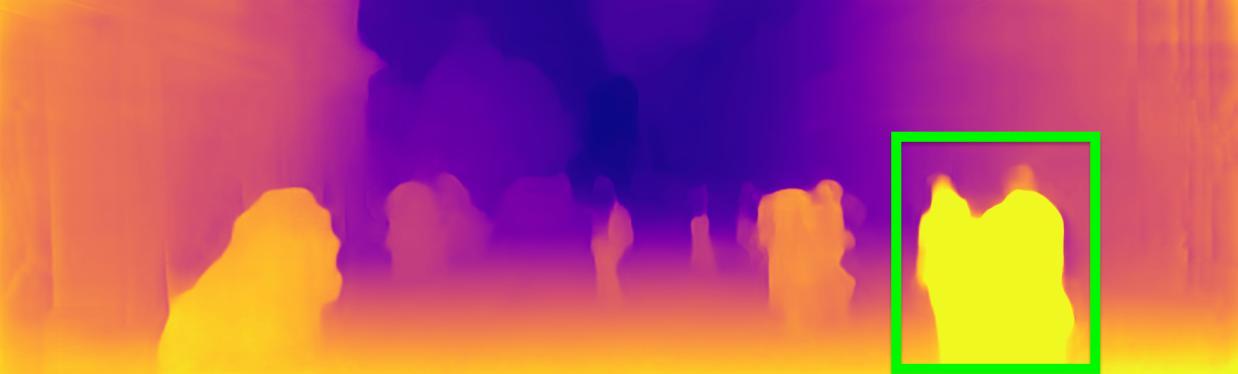}\includegraphics[  width=3.4cm,valign=t]{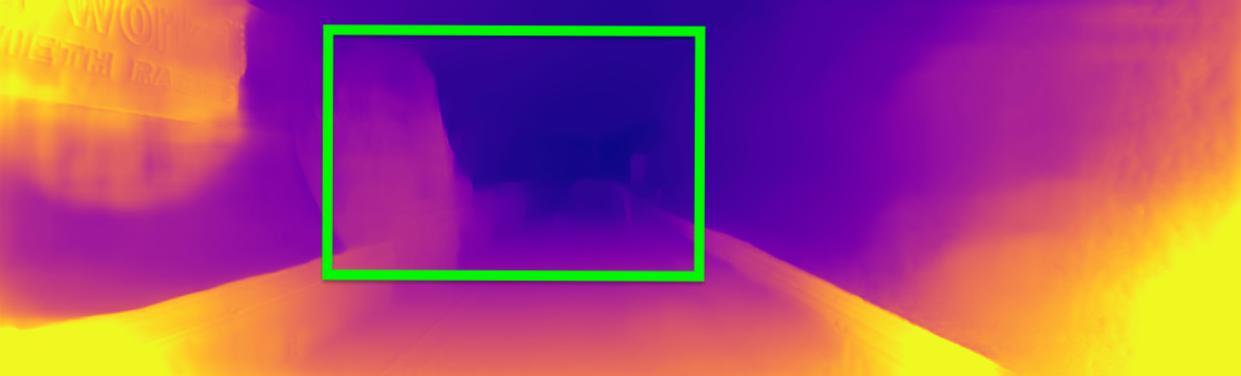}

\includegraphics[  width=3.4cm,valign=t]{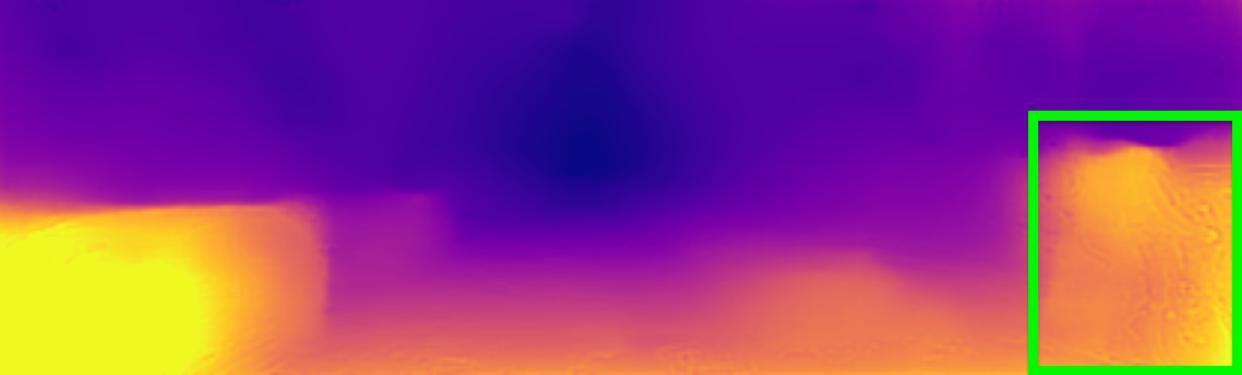}\includegraphics[  width=3.4cm,valign=t]{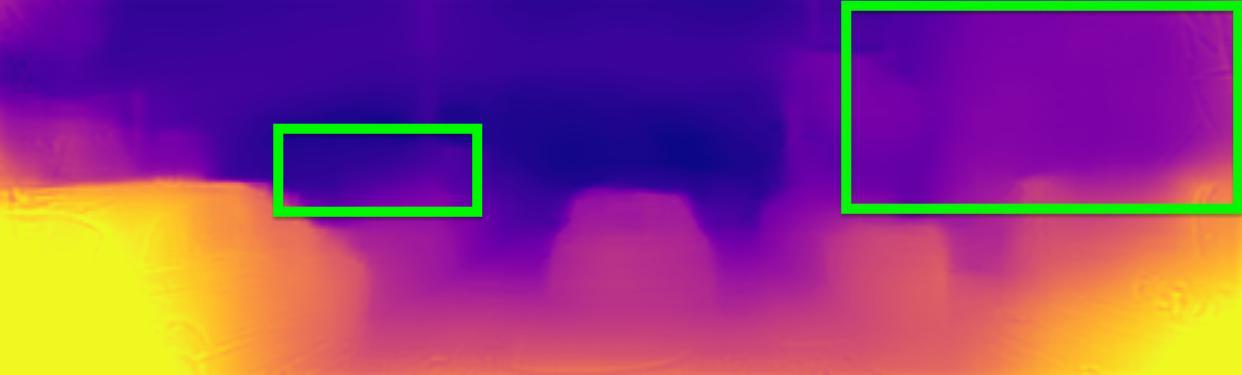}\includegraphics[  width=3.4cm,valign=t]{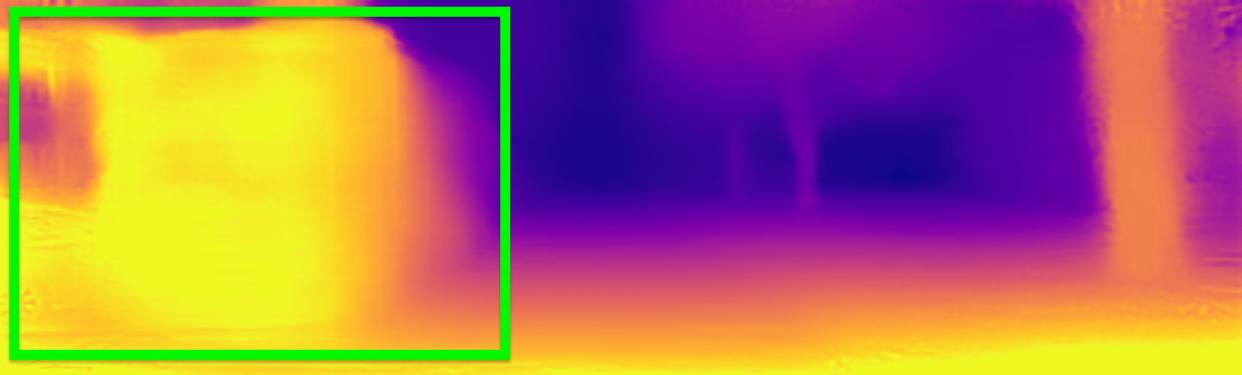}\includegraphics[  width=3.4cm,valign=t]{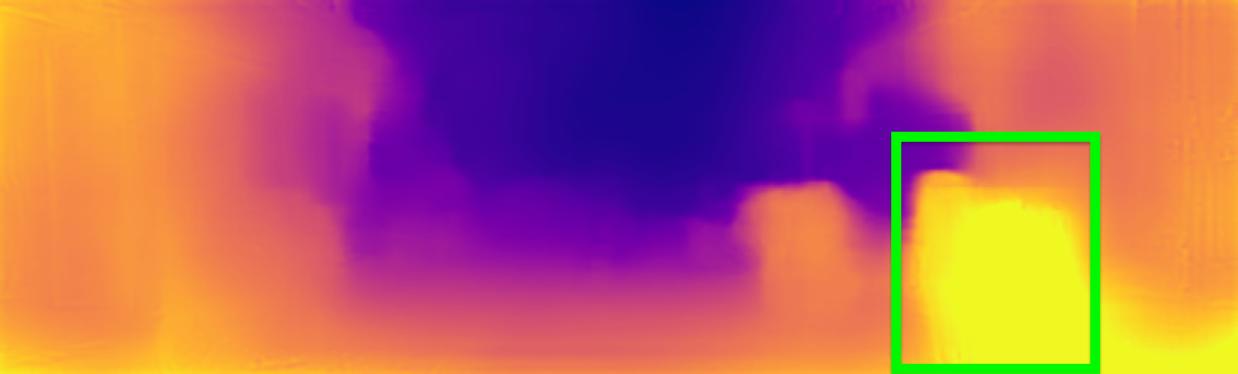}\includegraphics[  width=3.4cm,valign=t]{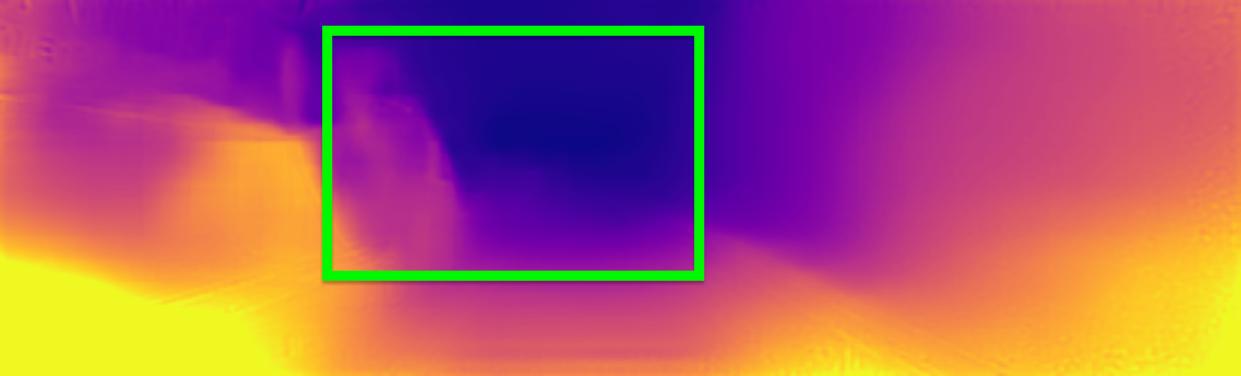}

\includegraphics[  width=3.4cm,valign=t]{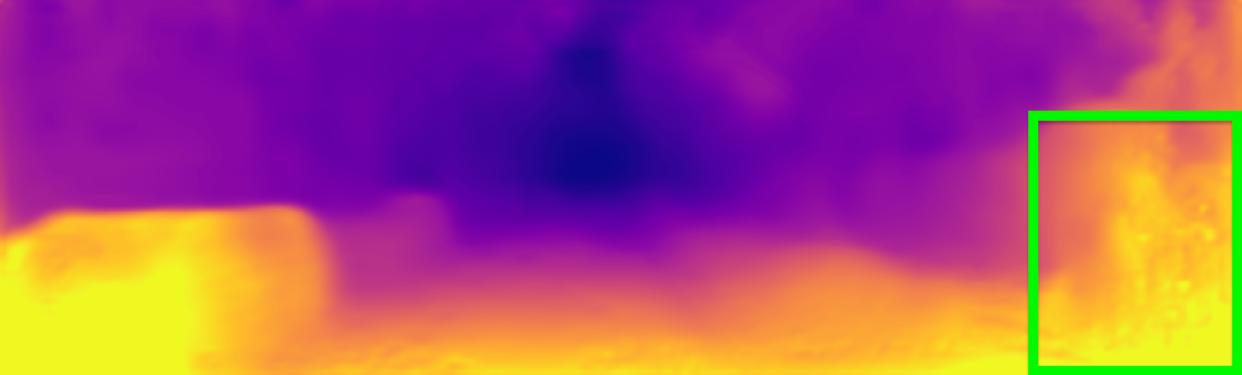}\includegraphics[  width=3.4cm,valign=t]{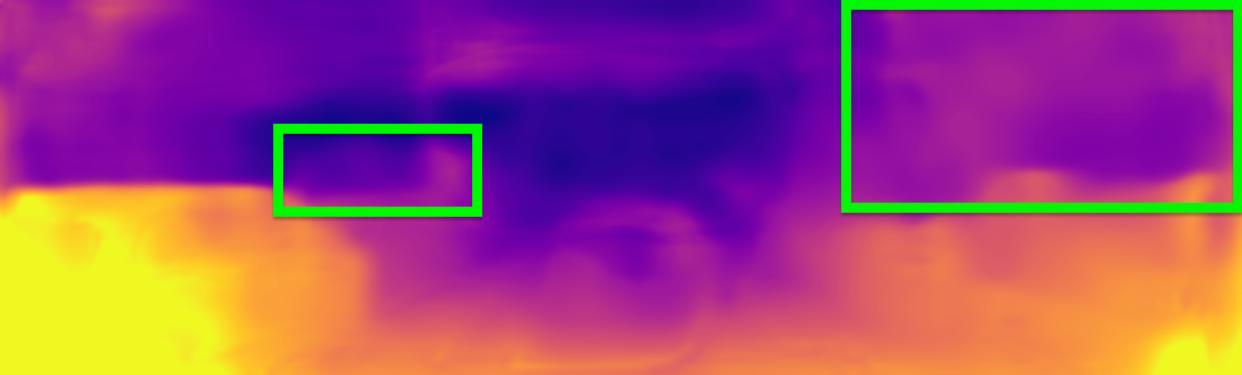}\includegraphics[  width=3.4cm,valign=t]{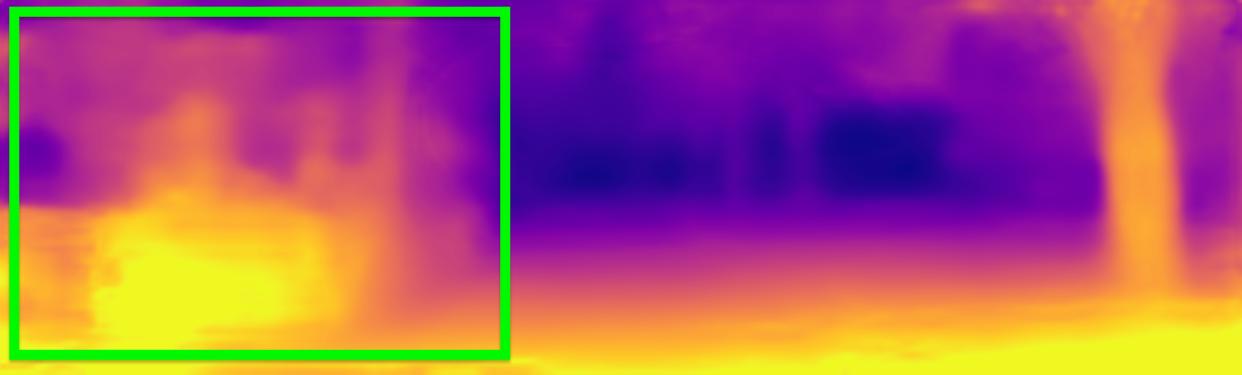}\includegraphics[  width=3.4cm,valign=t]{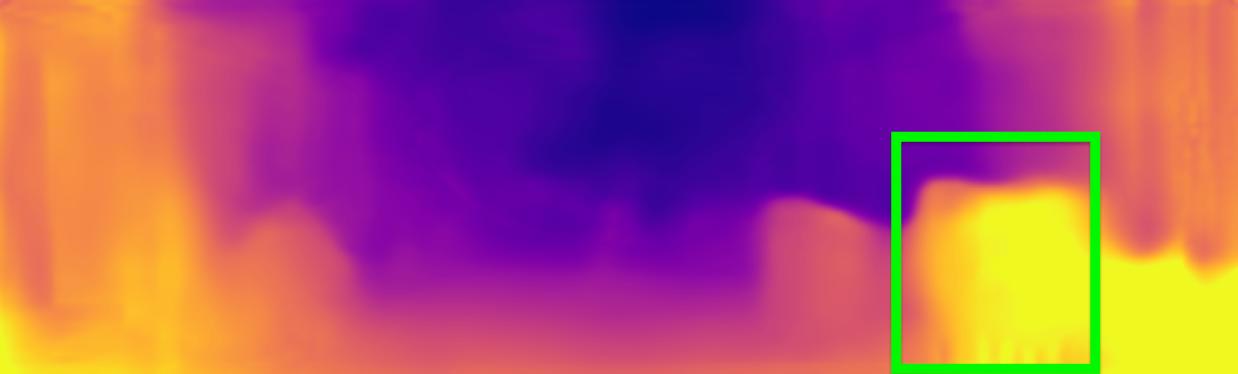}\includegraphics[  width=3.4cm,valign=t]{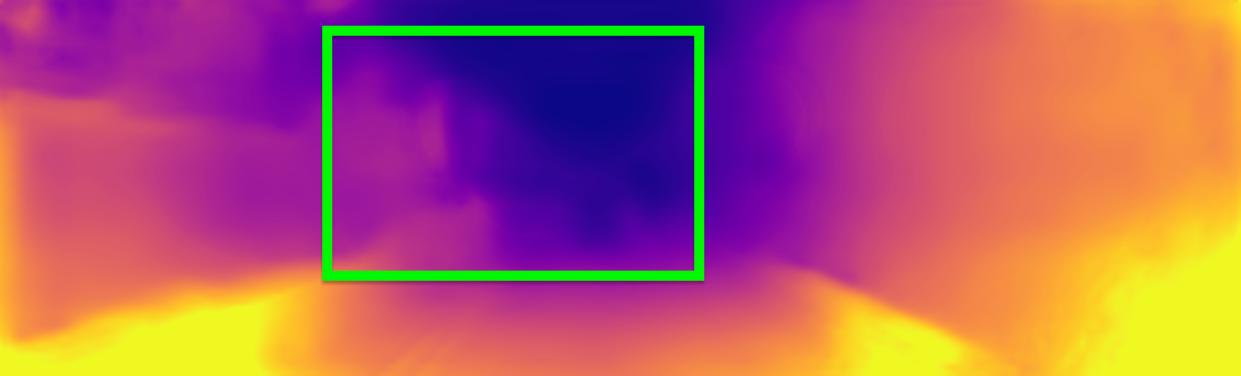}

\includegraphics[  width=3.4cm,valign=t]{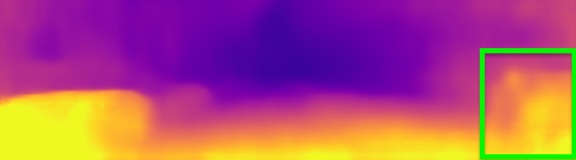}\includegraphics[  width=3.4cm,valign=t]{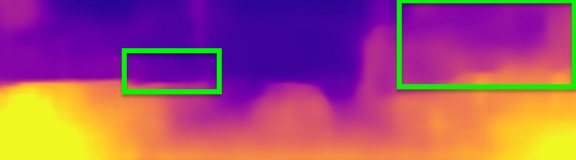}\includegraphics[  width=3.4cm,valign=t]{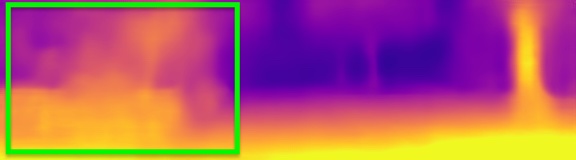}\includegraphics[  width=3.4cm,valign=t]{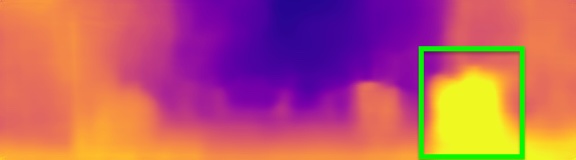}\includegraphics[  width=3.4cm,valign=t]{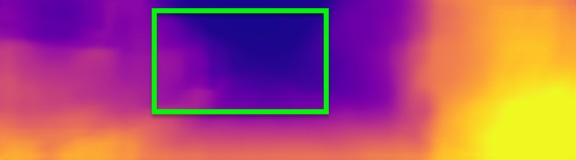}

\end{minipage}
\begin{minipage}[htbp]{0.005\textwidth}

\fontsize{7pt}{\baselineskip}\selectfont
\noindent

    \begin{turn}{90}{Df-Net~~   GeoNet~~~ DDVO~~~  MD2 ~~ \re{PackNet}~~\re{HR-Depth}~~\red{Feat(MS)}~MonoViT~~
    Ours~~~~~~~~\re{GT}~~~~ Input~~}
    \end{turn}
\end{minipage}
\end{center}
\caption{\red{Qualitative results on KITTI test set. Our method produces more accurate depth maps in low-texture regions, moving vehicles, and delicate structures. The advantages of our results are highlighted in green boxes. \red{Feat(MS) means the visualization results generated from FeatDepth \cite{shu2020featdepth} are using their models trained with both monocular and stereo inputs.}}}
\label{fig:inferresult}
\end{figure*}

\section{EXPERIMENTS}
\subsection{Network implementation}\label{networkarchi}
As shown in Figure \ref{fig:network}, our network is composed of three branches for offset learning, depth estimation, and pose estimation, respectively. The depth network adopts an encoder-decoder architecture in a U-shape with skip connections similar to DispNet \cite{mayer2016large}. The encoder takes a three-frame snippet as the sequential input, using the pre-trained Resnet \cite{he2016deep} as the backbone network. The depth decoder has three branches with shared weights, with a similar structure to \cite{godard2019digging}, using sigmoid activation functions in multi-scale side outputs and ELU nonlinear functions otherwise. The pose network takes two consecutive frames as input at each time and outputs the corresponding ego-motion, based on an encoder-decoder structure as well. For the DFA loss, we use a deformable alignment network to learn the feature alignment offset, also sharing weights with the deformable convolution used in the DepthNet branch for calculating the DFA Loss. \re{The three branches in the framework are jointly optimized during training while only DepthNet is used during inference.} Our models are implemented in PyTorch with the Adam optimizer in 4 Tesla V100 GPUs, using a learning rate $8e-5$ for the first 10 epochs and $8e-6$ for another 30 epochs. We trained models in monocular videos with a resolution of $640 \times 192$ (LR) and $1024 \times 320$ (HR) at a batch size of 4. 
\subsection{Evaluation metrics}
For real-world datasets, it is very hard to get ground truth depth. Most datasets use LIDAR sensors to scan the environment and utilize them as ground truth after being processed, for instance, KITTI \cite{geiger2012we} uses Velodyne laser scanner to collect the data. The most commonly used depth estimation benchmark of KITTI is the Eigen split, which is further improved by Zhou et al. \cite{zhou2017unsupervised} consisting of 39810 sequences for training, 4424 items for validation and 697 images for testing, respectively. And there are five evaluation metrics: \textbf{Abs Rel} for Absolute Relative Error, \textbf{Sq Rel} for Square Relative Error, \textbf{RMSE} for Root Mean Square Error, \textbf{RMSE log} for Root Mean Square Logarithmic Error and Accuracy:
\begin{itemize}
    \item $Abs~Rel = (1/n)\sum_{i\in n}((|d_i-d^{\ast}_i|)/d_i) $,
    \item $Sq~Rel = (1/n)\sum_{i\in n}((||d_i-d^{\ast}_i||^2)/d_i)$,
    \item $RMSE = ((1/n)\sum_{i\in n}||d_i-d^{\ast}_i||^2)^{1/2}$,
    \item $RMSE~log = ((1/n)\sum_{i\in n}||log(d_i)-log(d^{\ast}_i)||^2)^{1/2}$
    \item Accuracy: \% of $d_i$ s.t. $max((d_i/d^{\ast}_i), (d^{\ast}_i/d_i)) = \delta < \delta_n$,
\end{itemize}
where $n$ is the total number of pixels in the ground truth depth map, $d_i$ and $d^{\ast}_i$ represent the predicted and ground truth depth value of pixel $i$. $\delta_n$ denotes a threshold, which is usually set to $1.25^1$, $1.25^2$ and $1.25^3$.
\begin{table*}[t]

\begin{center}
\footnotesize
\caption{Quantitative performance of single depth estimation on KITTI eigen test set. For a fair comparison, all the results are evaluated at the maximum depth threshold of 80m. 
All methods are evaluated with raw LiDAR scan data. 
$\delta_1, \delta_2, \delta_3$ denote $\delta < 1.25, \delta < 1.25^2, \delta < 1.25^3$, respectively. The column ``train'' means training manners and ``M'' denotes self-supervised monocular training.}
\label{tab:backbone_resolution}
\setlength{\tabcolsep}{0.003\linewidth}
\begin{tabular}{ccccccccccc}
\toprule
\multirow{2}{*}{\!\!\!Methods\!\!\!}&\multirow{2}{*}{Train} &\multirow{2}{*}{Backbone}& \multirow{2}{*}{Resolution}&
 \multicolumn{4}{c}{Error metric$\downarrow$} & \multicolumn{3}{c}{Accuracy metric$\uparrow$} \\
\cmidrule(r){5-8} 
\cmidrule(r){9-11} 

  &&& &Abs Rel      & Sq Rel &   RMSE
&  RMSE log \!\!\!     &   {$\delta_1$}   &    {$\delta_2 $}
&   {$\delta_3$}       \\
\midrule

 {Ours (R18 LR)}	& {M}&ResNet18 & $640\times 192$		& 0.106	& 0.738	& 4.587	& 0.183	& 0.881	& 0.962	&  {0.983}\\
 {Ours (R50 LR)}	& {M}&ResNet50 & $640\times 192$		& 0.105	& 0.741	& 4.540	& 0.183	& 0.884	& 0.962	&  {0.983}\\
 {Ours (ViT LR)}	& {M}&ViT & $640\times 192$		&  {0.099}	&  {0.667}	&  {4.338}	& { 0.173}	&  {0.900}	&  {0.967}	& { 0.985}\\

 {Ours (R18 HR)}	& {M}&ResNet18	&$1024\times320$	& {0.103}	&0.764	&4.672	& {0.182}	&0.885&0.962	& {0.983}\\
 Ours (R50 HR)	& {M}&ResNet50	&$1024\times320$	& {0.102}	& {0.726}	& {4.479}	&0.179	& {0.889}	& {0.963}	& {0.983}\\
 {Ours (ViT HR)}	& {M}&ViT & $1024\times320$		&  {0.095}	&  {0.646}	&  {4.234}	& { 0.170}	&  {0.905}	&  {0.969}	& { 0.985}\\
\bottomrule

\end{tabular}
\end{center}
\end{table*}

\begin{table*}[htbp]
\small 
    \begin{center}
    \caption{Ablation study of the cross-view consistency loss on KITTI.}
    \label{tab:ablation}
    \begin{tabular}{l|c|c|c|c|c|c|c|c|c}
         
         \hline
         		&\!\!\!DFA \!\!\! &\!\!\!VDA\!\!\!	&\!\!\!Abs Rel\!\!\!	&\!\!\!Sq Rel\!\!\!	&RMSE	&\!\!\!RMSE log\!\!\!	&  $\!\!\delta \!< \!1.25$ \!\!  & \!\!  $\delta \!<\! 1.25^2 \!\! $
& \!\! $\delta \!<\! 1.25^3$ \!\!\!\\
       \hline
        Baseline	 	 		&	&	&0.124	&0.968	&5.030	&0.201	&0.855	&0.954	&0.980\\
         \hline
        Ours w/ DFA	 	 &$\surd$	&	&0.114	&0.873	&4.807	&0.191	&0.877	&0.960&0.982\\
        \hline
        Ours w/ VDA	  &	&$\surd$&0.110	&0.834	&4.694	&0.185	&0.885	&0.963	&0.983\\
        \hline
       \!\!\! Ours w/ DFA+VDA &$\surd$	&$\surd$&0.106	&0.738	&4.587	&0.183	&0.881	&0.962	&0.983\\
        \hline
\end{tabular}
\end{center}
\end{table*}
\subsection{Depth estimation evaluation}

We trained our models using the KITTI dataset \cite{geiger2012we}, which is the most commonly used benchmark in the field of depth estimation. During inference, we take only the reference frame as the input following the standard monocular test protocol proposed by Eigen \cite{eigen2014depth} and Zhou {\em et~al.} \cite{zhou2017unsupervised}.

The experimental results on the KITTI test set are presented in Table \ref{tab:1}. It is clear that the performance of our method outperforms prior SS-MDE methods in both the monocular and binocular training protocols. Some visual results are shown in Figure \ref{fig:inferresult}. As can be seen, our method can generate more accurate depth maps than other methods, especially in challenging cases, e.g., low-texture regions and moving objects (the moving cyclist in the first column and the moving cars in the last column).  {Although challenging cases usually only occupy a small portion of all scenes, it is worth noting that handling these challenging cases really matters for real-world applications like autonomous driving.}

\subsection{Ablation study and analysis}
\subsubsection{The impact of backbone and resolution}
We first ablate the performance of different backbone networks and input resolutions used in our method in Table \ref{tab:backbone_resolution}. The ViT-based network achieves better performance than the ResNet-based one due to its powerful representation ability. And the higher resolution tends to be  beneficial to lifting the performance.

\subsubsection{The effectiveness of DFA loss and VDA loss}
The following ablation study is conducted on KITTI using the most lightweight version (R18 LR) to highlight the effectiveness of the proposed two cross-view consistency losses. Table \ref{tab:ablation} shows the detailed results by adding specific loss to the baseline method. We trained the models of Monodepth2 \cite{godard2019digging} with resolution $640\times 192$ and batch size 4 as the baseline, taking the same setting as ours (R18 LR).

 \subsubsection{Ablation study and analysis of DFA loss}
\begin{figure*}[htbp]
\begin{minipage}{\textwidth}
\centering
\includegraphics[width=14cm]{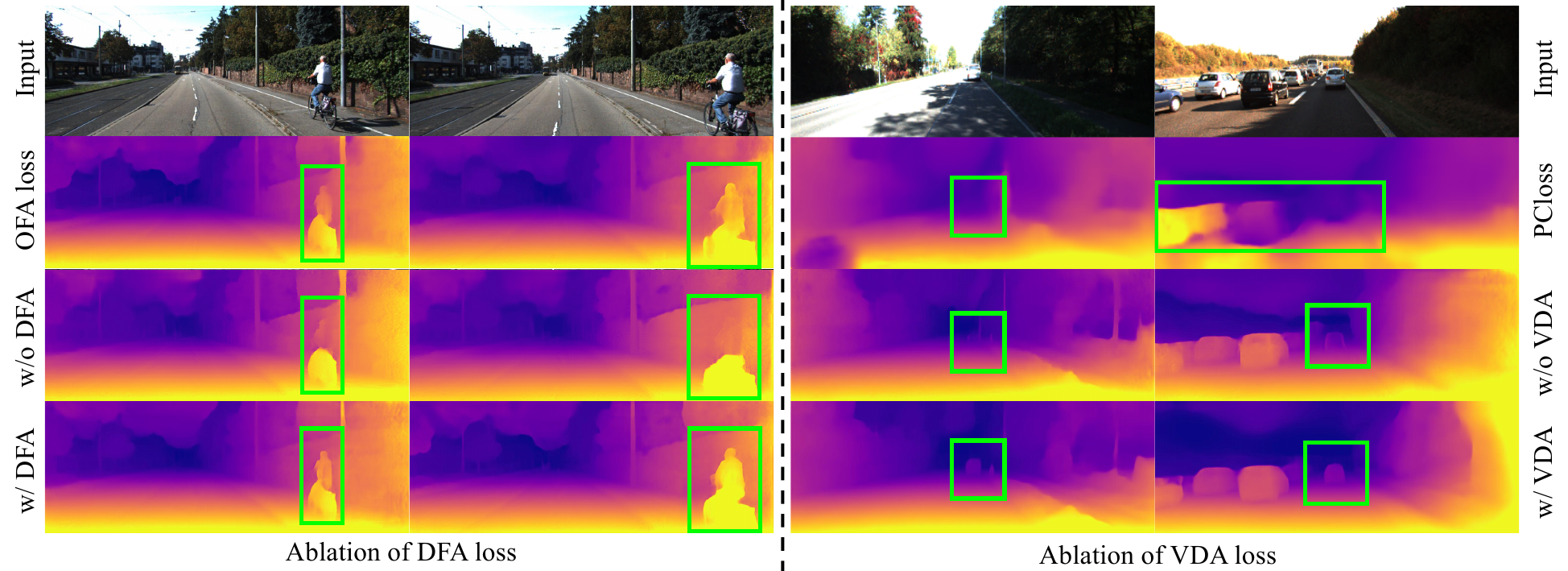}
\end{minipage}
\caption{{Visualization results of ablation study. \blue{The left shows the ablation of DFA loss and the comparison with the variant using OFA loss instead.
The right part shows the ablation of VDA loss and the comparison with the variant using PCloss instead.}}}
\label{fig:ablation}
\end{figure*}

 Our DFA loss aims to learn the temporal alignment from features of consecutive frames, which is used to guide temporal-consistent depth learning. We display samples of depth maps from one sequence in the left part of Figure \ref{fig:ablation}. As shown in the highlighted areas, our method can generate more accurate and coherent depth maps with sharper boundaries, especially in regions with illumination variance and low texture, e.g., the figure of the cyclist. The second and the third row show the results of the variants using optical flow alignment (more detail and analysis can be found in Section \ref{OFA}) and without any temporal alignment, respectively. 
 
Furthermore, experimental analysis of DFA loss is conducted, including the analysis of depth feature alignment offset, visualization of alignment offsets, and comparison experiment with alignment using optical flow.

\paragraph{\blue{Analysis of depth feature alignment offset}}

In our DFA loss, we use OffsetNet to learn the correspondences from deep features of consecutive RGB frames and use the learned correspondences to help to learn more coherent depth. In our understanding,
the feature offset is a kind of flow of features, which should share some similarities with optical flow. But they are region-wise rather than pixel-wise because the offsets are learned from deep features, which contain more semantic information and 3D geometry information. For the relationship between deformable convolution offset and optical flow, it is hard to clearly formulate. But researchers using deformable alignment in other areas \cite{chan2021understanding} are specifically dedicated to exploring this issue. They believe that deformable convolution can be decomposed into a combination of spatial warping and convolution. This decomposition can reveal the commonality of deformable alignment and flow-based alignment in formulation but with a key difference in their offset diversity. The offset diversity is closely related to the group number of offsets in the deformable convolutions. Their experiments demonstrate that the increased diversity in deformable alignment yields better aligned features, and hence significantly improves the quality of alignment. In our work, we set the group number of offset to $8$, therefore the deformable alignment learns $8\times 2\times 3 \times 3$, i.e., $144$ sets of offset for each pair of feature map to represent the temporal coherence of the features of continuous frames rather than only one offset learned in optical flow. We also use kernel size $5\times5$ to obtain a more diverse and higher-dimension ($8\times 2\times 5 \times 5$, i.e., $400$) offset field and conduct a comparison experiment. The experiment results show that the variant using a larger kernel size in DCN achieves similar results to the original version, i.e., Sq Rel: 0.738 vs. 0.735 and RMSE: 4.587 vs. 4.564 for $3\times3$ and $5\times 5$, respectively. We believe that the 144-dimension offset field is enough to enlarge the receptive field and model the temporal coherence of adjacent two frames. Taking model complexity into account, we choose kernel size $3\times 3$ in our final version.

\paragraph{\blue{Visualization of depth feature alignment offset}}
\begin{figure}[t]
  \begin{center}
  \begin{minipage}{0.8\linewidth} 
   \includegraphics[width=\textwidth]{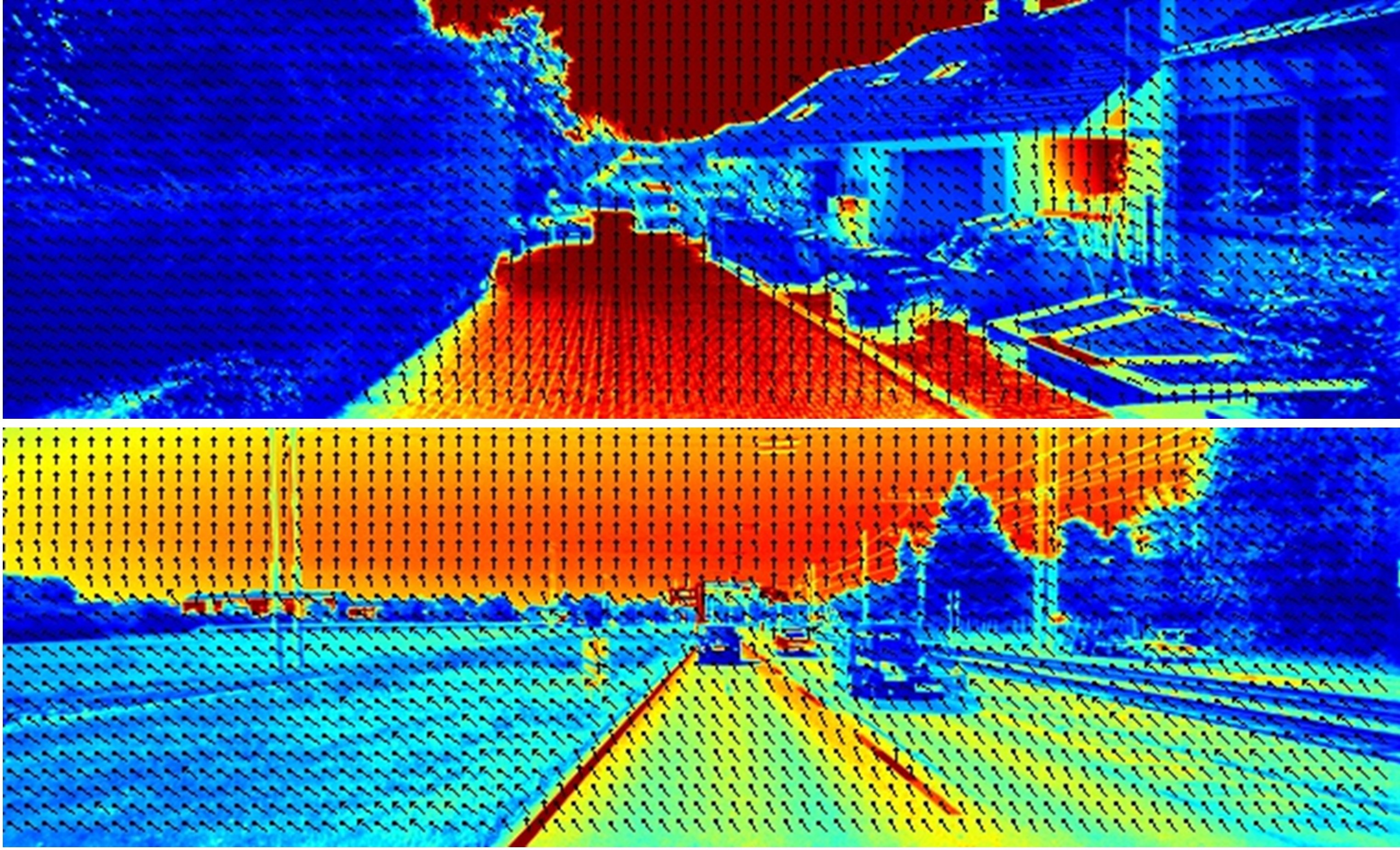}
  \caption{Samples of the visualization of the learned depth feature offset.}
  \label{offsetfigure}
  \end{minipage}
   \end{center}
\end{figure}
We adopt PCA decomposition to select one set offset to visualize the heatmaps of learned feature alignment offset in Figure \ref{offsetfigure}. Hotter colors denote higher values. Our OffsetNet pays more attention to stable regions and semantic outlines such as ground and wall to learn useful alignment offset. Moreover, the direction of offsets in the same semantic part tends to be the same, which implies the alignment offsets are learned in a ``region-to-region'' manner instead of varying with pixels.
\paragraph{\blue{Compared with alignment using optical flow}}\label{OFA}
To demonstrate the effectiveness of the DFA loss, we replace the DFA loss with the optical flow of the corresponding feature to conduct the depth feature alignment. The evaluation result is shown in Table \ref{OFAloss}, where the ``OFA loss" denotes the variant using optical flow alignment. The evaluation results show that using optical flow as a single channel offset to regularize the depth
learning turns out to be not feasible, which verifies the effectiveness of our DFA loss.

\begin{table*}[htbp]
\footnotesize
    \begin{center}
    \caption{ {Experiment results compared with the variant using optical flow alignment.}}
    \label{OFAloss}
    \setlength{\tabcolsep}{0.012\linewidth}
    \begin{tabular}{lccccccc}
         \toprule
         		Methods&Abs Rel	&Sq Rel	&RMSE	&RMSE log	&  $\delta < 1.25$   &   $\delta < 1.25^2 $
&  $\delta < 1.25^3$ \\
        \midrule
        Ours (DFA loss)			& 0.106	& 0.738	& 4.587	& 0.183	& 0.881	& 0.962	& 0.983\\
        \midrule
        Ours (OFA loss)&0.119	& 0.902	& 4.860	& 0.194	& 0.863	& 0.958	& 0.982\\
        
         \midrule
\end{tabular}
\end{center}
\end{table*}

\subsubsection{Ablation study and analysis of VDA loss}
Our VDA loss is designed to exploit temporal coherence in voxel space to ensure the models' robustness and tolerance to challenging cases, especially moving objects. Samples of depth images produced by models with and without VDA loss are shown in Figure \ref{fig:ablation}. The right parts show the superiority of our method in handling moving objects. The depth maps in the last row (with VDA loss) are of higher quality than those in the third row (without VDA loss) and the second-row PCloss (baseline method with Point cloud alignment loss \cite{mahjourian2018unsupervised}), confirming the effectiveness of VDA loss in moving objects regions.

Besides, we further analyse VDA loss including the analysis of VDA loss in handling moving objects and the analysis of hyper-parameters in VDA loss.

\paragraph{\blue{The effectiveness of VDA loss in handling moving objects}} We conducted quantitative and qualitative experiments to validate the effectiveness of our VDA loss. We split the whole test set into two parts namely motion set and static set, respectively, according to if there is a moving object in the scenes. We evaluated our method and other representative methods in both sets. The results are reported in Table \ref{tab:motion}. \blue{According to evaluation results, our method achieves the best performance in the motion test set. For the static test set, our method performs the best in three error metrics of four and the second best in the accuracy metrics with marginal gaps (0.002, 0.001, and 0.001, respectively). It is noted that the loss is proposed to keep the method performing stably towards violation factors. According to the experiment results, our method achieves a better trade-off than other methods in the two settings.} For the quantitative comparison, it is clear that our method is more robust to scenes with object motions than other methods. Moreover, we show several visual samples in Figure \ref{fig:ablation}. As shown in the right part of Figure \ref{fig:ablation}, our method can robustly and clearly infer the moving vehicles in outdoor driving scenes, while other methods may lose or misestimate the moving objects.

\paragraph{\blue{Analysis of \blue{point cloud partitioning methods} in VDA loss}}

Our VDA loss exploits cross-view consistency of the distribution of point clouds in voxel space. It is interesting to investigate the impact of the way of dividing the 3D space into voxels, i.e., the number of voxels along each axis $N_x, N_y, N_z$ \blue{and the partitioning methods}. Therefore, we conduct an ablation study of different numbers of voxels, reported in Table \ref{voxeltab}. The size of the voxel has a close relation to the tolerance to moving objects. According to our prior knowledge, the objects moving in the vertical direction are almost impossible on the outdoor driving dataset. We thus set  {$N_z$} to 24 and evaluate the different sizes of the voxel by changing $N_x$ and $N_z$. As shown in Table \ref{voxeltab}, the number of voxels along axis $x$ and axis $z$ can slightly affect the performance. Assuming a border case, if $N_x= N_y=N_z=1$, which means regarding the whole space as one voxel, the voxel density is the same and the VDA loss always equals 0. However, if the values of $N_x, N_y, N_z$ are too large, the voxel size will be very small, which is contrary to the goal of VDA loss. Therefore, large values for $N_x, N_y, N_z$ are not recommended.
\blue{Besides, we conduct an experiment to investigate the impact of voxel partitioning methods. The experiment involved employing the SID strategy from the work \cite{fu2018deep} for point cloud discretization, which utilizes a log-space discretization of the depth dimension, thus down-weighing the training losses in regions with large depth values. According to the results in Table \ref{voxeltab}, the overall performance of both the two division manners is comparable. We speculate that this is because voxel density loss is not sensitive to voxel partitioning methods, as long as the partitioning methods of adjacent frames are consistent.}
\begin{table*}[htbp]
\footnotesize 
    \begin{center}
    \caption{Evaluation results on the split motion and static test sets.}
    \label{tab:motion}
    \setlength{\tabcolsep}{0.003\linewidth}
    \begin{tabular}{l|c|c|c|c|c|c|c|c}
         \hline
         		&Region	&Abs Rel	&Sq Rel	&RMSE	&RMSE log	&  $\delta < 1.25$   &   $\delta < 1.25^2 $
&  $\delta < 1.25^3$ \\
        \hline
        Baseline \cite{godard2019digging} (R18 LR) &motion&0.118	&0.978	&5.079	&0.199&0.874&0.957&0.980\\
        \hline
        FeatDepth \cite{shu2020featdepth} (R50 HR)&motion& {0.108}	& {0.987}	& {4.910}	& {0.189}& {0.877}& {0.958}& {0.981}\\
        \hline
         Ours (R18 LR) &motion& 0.106	& 0.790	& 4.769	& 0.186	& 0.882& 0.959& 0.982\\
         \hline
        Ours (R50 HR) &motion&  {\textbf{0.102}}	&  {\textbf{0.734}}	&  {\textbf{4.434}}	&  {\textbf{0.179}}	&  {\textbf{0.889}}&  {\textbf{0.963}}&  {\textbf{0.983}}\\
        \hline
        Baseline \cite{godard2019digging} (R18 LR) &static&0.109	&0.704	&3.999	&0.173	&0.888	&\textbf{0.968}	&\textbf{0.986}\\
        \hline
        FeatDepth \cite{shu2020featdepth} (R50 HR)&static& {0.104}	& {0.632}	& {\textbf{4.079}}	& {0.173}& {\textbf{0.892}}& {0.967}& {0.985}\\
        \hline
         Ours (R18 LR) &static & 0.106	& 0.641	& 4.250	& 0.177	& 0.879	& 0.967	& 0.985\\
         \hline
          Ours (R50 HR) &static &  {\textbf{0.103}}	&  {\textbf{0.627}}	&  {4.136}	& {\textbf{ 0.173}}	&  {0.890}	&  {0.967}	&  {0.985}\\
         \hline
\end{tabular}
\end{center}
\end{table*}{}
        
\begin{table*}[htbp]
\footnotesize
    \begin{center}
    \caption{{Experiment results for \blue{the analysis of partitioning methods in VDA loss.} $N_x, N_y, N_z$ denote the number of voxels divided along three axes.}}
    \label{voxeltab}
    \setlength{\tabcolsep}{0.007\linewidth}
    \begin{tabular}{ccccccccc}
         \toprule
         	\blue{Partitioning}	& ($N_x, N_y, N_z$)&Abs Rel	&Sq Rel	&RMSE	&RMSE log	&  $\delta < 1.25$   &   $\delta < 1.25^2 $
&  $\delta < 1.25^3$ \\
        \midrule
       \blue{Uniform}& $(20,20,24)$&0.108  &   0.855  &   4.912  &   0.183  &   0.882  &   0.961  &   0.983\\
        \midrule
        \blue{Uniform}&$(40,40,24)$&0.106	& 0.738	& 4.587	& 0.183	& 0.881	& 0.962	& 0.983\\
        \midrule
       \blue{ Uniform}&$(60,60,24)$ &   0.107  &   0.737  &   4.742  &   0.184  &   0.875  &   0.960  &   0.983\\
       \midrule
        \blue{SID \cite{fu2018deep}} &\blue{$(40,40,24)$}&   \blue{0.105}  &  \blue{0.736}  &   \blue{4.743}  &   \blue{0.183} &   \blue{0.877}  &   \blue{0.961}  &   \blue{0.983}\\
         \bottomrule
\end{tabular}
\end{center}
\end{table*}{}

\begin{table}[t]
\begin{minipage}{\linewidth}
\footnotesize
\caption{Test results on Make3D.  {Column ``Train" with label ``M"/``S" means training with monocular/stereo data. }}
\label{maketabel}
\begin{center}
\begin{tabular}{lccccc}
\toprule
\multirow{1}{*}{Methods}& \multirow{1}{*}{Train}& 
Abs Rel      & Sq Rel &   RMSE
&  log10       \\
\midrule
 {Monodepth \cite{godard2017unsupervised}}	& {S}	& {0.544}	& {10.94}	& {11.760}	& {0.193}\\
SFMlearner \cite{zhou2017unsupervised}	&M	&0.383	&5.321	&10.470	&0.478\\
DDVO \cite{wang2018learning}	&M	&0.387	&4.720	&8.090	&0.204\\
Monodepth2 \cite{godard2019digging} 	&M	&0.322	&3.589	&7.417	&0.163\\
 
 {Johonston et al. \cite{johnston2020self}}& {M}& {0.297}	& { {2.902}}	& {7.013}	& {0.158}\\
 {FeatDepth \cite{shu2020featdepth}}& {M}& {0.313}	& {3.489}	& {7.228}	& {0.158}\\
 Ours (R18 LR)	&M	& 0.316	& 3.200	& 7.095	& 0.158\\
 {Ours (R50 HR)}	&{M}	& {0.290}	& {3.070}	& {6.902}	& {0.155}\\
 MonoViT 	&M	& 0.286	& 2.758	& 6.623	&0.147\\
Ours (ViT HR) &M	& 0.283	& 3.021	& 6.579	&0.145\\
\bottomrule
\end{tabular} 
\end{center}
\end{minipage}

\end{table}
\begin{figure}

\begin{center}
\begin{minipage}{\linewidth}
\centering
\includegraphics[width=\textwidth]{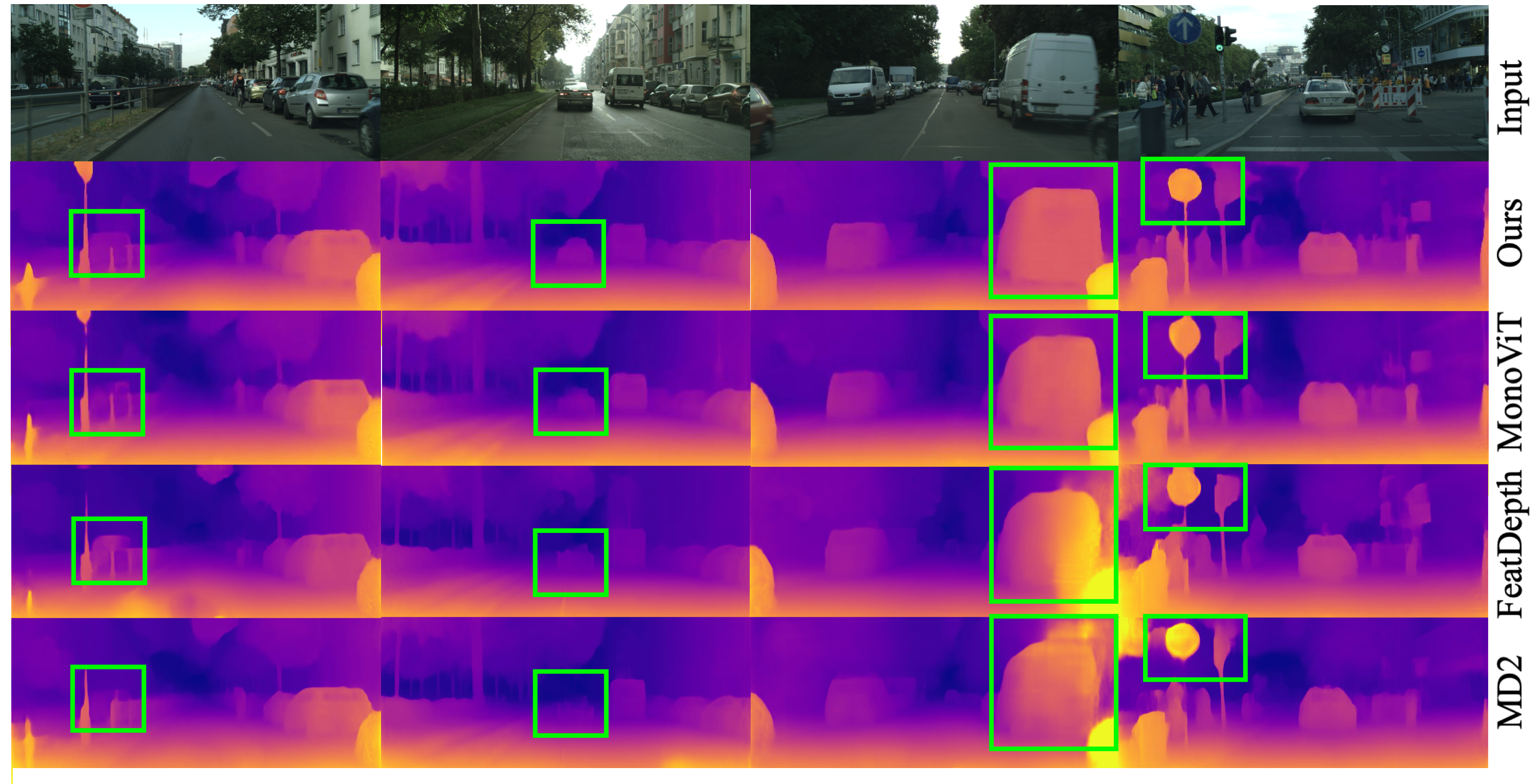}

\end{minipage}

\end{center}
\caption{Qualitative results on Cityscapes \cite{cordts2016cityscapes}. Our method produces more accurate depth maps in moving objects and texture-less regions. {Green boxes highlight the difference between the predicted depth maps by different methods.}}
\label{city9}
\end{figure}

\subsection{Evaluation of generalization ability}\label{generalization}
Though our models were only trained on KITTI \cite{geiger2012we}, competitive results can be achieved on unseen datasets without any fine-tuning. {We display the generalization ability of our method using the version ours (R18 LR).}
\begin{figure}
\begin{center}
\begin{minipage}{\linewidth}
\centering
\includegraphics[width=\textwidth]{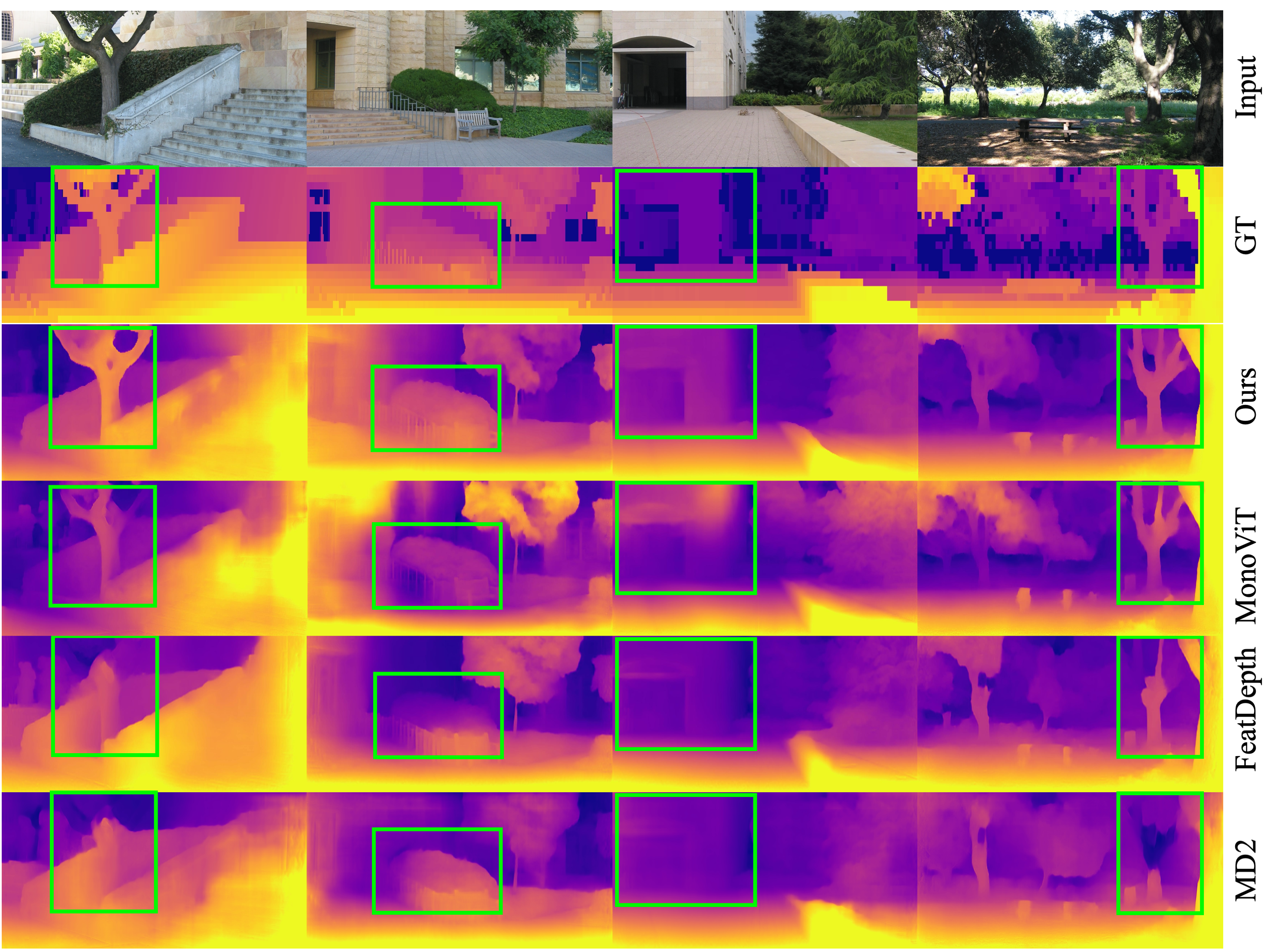}

\end{minipage}


\end{center}
\caption{Qualitative results on Make3D \cite{saxena2008make3d}. Our method can generate more accurate depth maps. {Green boxes highlight the difference between the predicted depth maps by different methods.}}

\label{fig:make}
\end{figure}

{\bf Cityscape.} The challenges of Cityscape \cite{cordts2016cityscapes} mainly arise from the poor lighting condition, raining weather, and moving objects. We cropped out the bottom part of the original images to remove the car hoods. The generated depth maps show the good domain adaptation ability of our models. Compared with other state-of-the-art approaches {Monodepth2 \cite{godard2019digging}, FeatDepth \cite{shu2020featdepth}} and MonoViT \cite{monovit}, our method is more accurate at perceiving moving or distant objects and delicate structures, as shown in Figure \ref{city9}{, especially the part highlighted in the green boxes.}

{\bf Make3D.} We conducted the center cropping and scaling \cite{godard2019digging} for the alignment between input images and the ground truth \cite{saxena2008make3d}. In Table \ref{maketabel}, our results with different settings outperform the baseline methods Monodepth2 \cite{godard2019digging} and MonoViT \cite{monovit}. The qualitative comparison in Figure \ref{fig:make} provides additional intuitive evidence on the generalization ability of our method. 
\begin{table}[t]
\centering
\begin{minipage}{\linewidth}  
\footnotesize
\caption{Comparison of model complexity.}\label{inferencetable}
\footnotesize
\begin{center}
\begin{tabular}{lccc}

         \toprule
         			&Params (M)$\downarrow$	&FLOPs (G)$\downarrow$	&FPS$\uparrow$	\\
        \midrule
        Monodepth2 \cite{godard2019digging}&14.33	&8.03	&184.5	\\
        Ours (R18 LR) &14.33	&8.03	&184.5	\\
        FeatDepth \cite{shu2020featdepth}&33.16	&85.11	&	59.6\\
        {Ours (R50 HR)} &{32.52}	& {44.31}	& {87.3}	\\
         MonoViT \cite{monovit} &27&60.25 &18\\
         Ours (ViT LR) &27& 60.25&18\\
         \bottomrule

\end{tabular}
\end{center}
\end{minipage}
\end{table}
\subsection{Model complexity analysis}
Since depth estimation methods are often used in autonomous driving or drone systems, model size, and speed during inference are very important. Therefore, we compare our model with prior works in terms of parameters (M), computations (FLOPs), and inference speed (FPS), as shown in Table \ref{inferencetable}. Our method proposes two cross-view loss items to regularize the temporal coherence during training, which are not used during inference. Therefore, the model complexity of our methods is consistent with our baseline method.

\section{Conclusion and discussion}\label{conclusion}


This study is dedicated to the SS-MDE problem with a focus on robust cross-view consistency. We first propose DFA loss to exploit the temporal coherence in feature space to produce consistent depth estimation. Compared with the photometric loss in the RGB space, measuring the cross-view consistency in the depth feature space is more robust in challenging cases such as illumination variance and texture-less regions, owing to the representation power of deep features. Moreover, we design VDA loss to exploit robust cross-view 3D geometry consistency by aligning point cloud distribution in the voxel space.  {VDA loss has shown to be more effective} in handling moving objects and occlusion regions than the rigid point cloud alignment loss. Experimental results  {on outdoor benchmarks} demonstrate that our method has achieved superior results than state-of-the-art approaches and can generate better depth maps in texture-less regions and moving object areas.  {More efforts can be made to improve the voxelization method in VDA loss to enhance the generalization ability and apply the proposed method to indoor scenes, which will be left for future work.}

\bibliographystyle{IEEEtran}
\bibliography{reference}

\vfill

\end{document}